\documentclass[11pt]{article}

\usepackage[final]{acl}

\usepackage{times}
\usepackage{latexsym}
\usepackage{adjustbox}
\usepackage{enumitem}
\usepackage{booktabs} 
\usepackage{footnote}
\usepackage{multicol}
\usepackage{arydshln}
\usepackage[T1]{fontenc}

\usepackage[utf8]{inputenc}

\usepackage{microtype}
\usepackage{colortbl}
\usepackage{inconsolata}

\usepackage{graphicx}
\usepackage{multirow}
\usepackage{float}
\usepackage{array}
\usepackage{amssymb} 
\usepackage{pifont}
\usepackage{amsmath}
\usepackage{makecell} 
\usepackage{tabularx} 
\definecolor{darkred}{RGB}{190,0,0}
\definecolor{darkgreen}{RGB}{84,135,49}

\newcommand{\best}[1]{\textbf{#1}} 
\newcommand{\MinSup}{\mathrm{MinSup}}
\title{LogicGraph : Benchmarking Multi-Path Logical Reasoning via Neuro-Symbolic Generation and Verification}

\author{Yanrui Wu$^{1,4}$,\; \textbf{Lingling Zhang}$^{1,4}$ \thanks{Corresponding author}, \; Xinyu Zhang$^{1,4}$,\; Jiayu Chang$^{2}$,\; Pengyu Li$^{1,4}$,\; \\ \textbf{Xu Jiang}$^{3}$,\; \textbf{Jingtao Hu}$^{1}$,\; \textbf{Jun Liu}$^{1,5}$
\\
$^{1}$School of Computer Science and Technology, Xi'an Jiaotong University \; 
\\
$^{2}$Department of Electrical Engineering, Stanford University \; 
\\
$^{3}$School of Computer Science and Technology, Tiangong University \; \\
$^{4}$Ministry of Education Key Laboratory of Intelligent Networks and Network Security, China \; \\
$^{5}$Shaanxi Province Key Laboratory of Big Data Knowledge Engineering, China\; \\
\texttt{{yanrui.wu}@stu.xjtu.edu.cn, zhanglling@xjtu.edu.cn}
}

\begin{document}
\maketitle
\vspace{10pt} 
\begin{abstract}
Evaluations of large language models (LLMs) primarily emphasize convergent logical reasoning, where success is defined by producing a single correct proof. However, many real-world reasoning problems admit multiple valid derivations, requiring models to explore diverse logical paths rather than committing to one route. To address this limitation, we introduce \textbf{LogicGraph}, the first benchmark aimed to systematically evaluate multi-path logical reasoning, constructed via a neuro-symbolic framework that leverages backward logic generation and semantic instantiation. This pipeline yields solver-verified reasoning problems formalized by high-depth multi-path reasoning and inherent logical distractions, where each instance is associated with an exhaustive set of minimal proofs. We further propose a reference-free evaluation framework to rigorously assess model performance in both convergent and divergent regimes. Experiments on state-of-the-art language models reveal a common limitation: models tend to commit early to a single route and fail to explore alternatives, and the coverage gap grows substantially with reasoning depth. LogicGraph exposes this divergence gap and provides actionable insights to motivate future improvements. Our code and data will be released at \url{https://github.com/kkkkarry/LogicGraph}.

\end{abstract}

\section{Introduction}
Logical reasoning is central to general intelligence, enabling systems to derive valid conclusions from given premises~\cite{kaufman2011general}. However, most existing benchmarks for evaluating LLM reasoning~\cite{han2024folio, xu2025large, huang2025math, fu2025geolaux} primarily emphasize \textit{convergent} thinking---assessing whether a model can reach a correct final conclusion for a given problem. This emphasis leaves underexplored what Guilford terms \textit{divergent} thinking~\cite{guilford1967nature}: the ability to actively generate multiple plausible alternatives for the same conclusion in a problem.

\begin{figure}[t]
    \centering
    \includegraphics[width=\columnwidth]{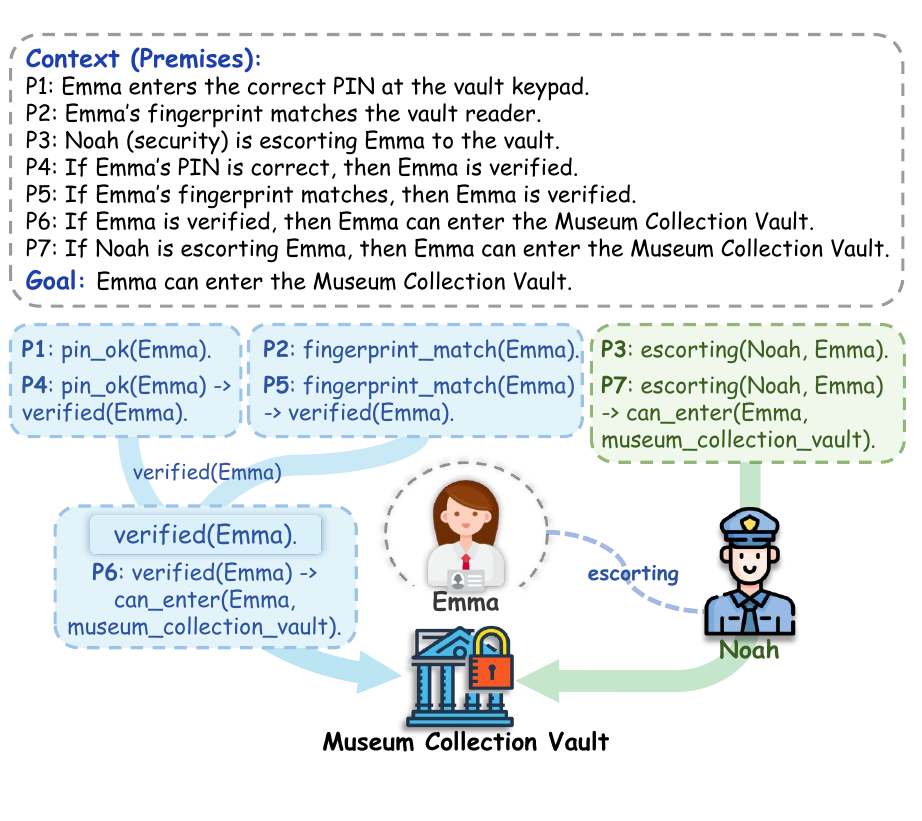}
    \vspace{-25pt}
    \caption{Illustration of the multi-path reasoning challenge. In the real world, the same conclusion may be entailed via multiple derivation paths.}
    \label{fig:example}
    \vspace{-5px}
\end{figure}

\begin{table*}
  \centering
  \begin{adjustbox}{width=\textwidth}
  \begin{tabular}{lccccccccc}
  \toprule
    \textbf{Benchmark} &
    \makecell{\textbf{Symbolic} \\ \textbf{Notation}} &
    \textbf{Distraction} &
    \makecell{\textbf{Depth} \\ \textbf{(avg.)}} &
    \makecell{\textbf{Paths} \\ \textbf{(range)}} &
    \makecell{\textbf{Reuse Ratio} \\ \textbf{(range)}} &
    \makecell{\textbf{Task Type}} &
    \makecell{\textbf{Stepwise} \\ \textbf{Evaluation}} & \\
  \midrule
    ProofWriter   
        & {\color{darkred}\ding{55}}    
        & {\color{darkred}\ding{55}}
        & 2.2   & 1 & 1.0   & Binary & {\color{darkred}\ding{55}}\\
    FOLIO
        & {\color{darkgreen}\ding{51}}
        & {\color{darkred}\ding{55}}
        & 3.4   & 1   & 1.0  & Binary & {\color{darkred}\ding{55}}\\
    ProntoQA 
        & {\color{darkgreen}\ding{51}}
        & {\color{darkred}\ding{55}}
        & --    & --   & -- & Binary & {\color{darkred}\ding{55}}\\
    RuleTaker
        & {\color{darkred}\ding{55}}    
        & {\color{darkred}\ding{55}}
        & 2.4   & 1  & 1.0  & Binary & {\color{darkred}\ding{55}}\\
    LogicBench
        & {\color{darkred}\ding{55}}
        & {\color{darkred}\ding{55}}    
        & --    & --   & --  & Binary/MCQA
        & {\color{darkred}\ding{55}}\\
    LogicNLI
        & {\color{darkred}\ding{55}}
        & {\color{darkred}\ding{55}}    
        & --    & --   & -- & Ternary & {\color{darkred}\ding{55}}\\
    Multi-LogiEval
        & {\color{darkred}\ding{55}}  
        & {\color{darkred}\ding{55}}
        & 2.51   & 1  & 1.0  & Binary & {\color{darkred}\ding{55}}\\
    ProverQA
        & {\color{darkgreen}\ding{51}}  
        & {\color{darkgreen}\ding{51}}
        & 4.73  & 1   & 1.0  & Ternary & {\color{darkred}\ding{55}}\\
  \midrule
    \textbf{LogicGraph}
        & {\color{darkgreen}\ding{51}}  
        & {\color{darkgreen}\ding{51}}
        & \best{6.01} & \best{2 $\sim$19}
        & \best{1.0 $\sim$1.9}  & \textbf{Proof Gen.}
        & {\color{darkgreen}\ding{51}}\\
    \bottomrule
  \end{tabular}
  \end{adjustbox}
   \caption{Comparison of LogicGraph (ours) with existing logical reasoning datasets across key characteristics. Distraction refers to the simultaneous occurrence of logical and semantic distraction.}
  \label{tab:comparison}
  \vspace{-5px}
\end{table*}

In real-world scenarios, reasoning is rarely linear; agents may reach the same conclusion through multiple valid lines of argument and often benefit from exploring alternatives~\cite{yao2023tree, besta2024graph}. Consider the access-control scenario in Figure~\ref{fig:example}. The conclusion ``Emma can enter the Vault'' can be entailed via distinct mechanisms, e.g., through PIN verification (P1, P4, P6), biometrics (P2, P5, P6), or a security escort (P3, P7), illustrating that multiple derivation paths may support the same claim. To bridge this gap between real-world reasoning demands and current benchmark objectives, we formulate the multi-path logical reasoning task: given premises and a conclusion, the goal is to enumerate the distinct logically valid derivation paths that entail the conclusion, rather than merely outputting a binary label.

However, establishing a benchmark for multi-path reasoning faces three fundamental challenges: \textit{(i) Scalable Construction} of exhaustive ground-truth derivation paths; \textit{(ii) Reliable Evaluation} of open-ended generation; and \textit{(iii) Cognitive Assessment} of exploration versus mere correctness. To address these challenges, we introduce LogicGraph, the first benchmark and evaluation framework designed for multi-path logical reasoning.

\textbf{Construction via Reverse Logic DAGs.}
Manually annotating every valid reasoning path in complex text is prohibitively error-prone. To guarantee exhaustive ground truth, we employ a backward construction paradigm~\cite{al2015comparison}. We synthesize premises from the conclusion upwards using fundamental argument forms, creating a symbolic logic Directed Acyclic Graph (DAG) that is subsequently verbalized. This DAG-based construction yields three characteristics of LogicGraph: high-depth multi-path reasoning (with 2--19 valid paths per query and an average depth of 6.01), inference node reuse, and inherent logical distractions. See Table~\ref{tab:comparison} for more details.

\textbf{Neuro-Symbolic Evaluation.}
Evaluating divergent outputs is difficult because traditional string matching is brittle and LLM-as-a-Judge approaches suffer from hallucination~\citep{wang2025trustjudge, szymanski2025limitations}. We propose a \textit{reference-free neuro-symbolic} evaluator that translates generated natural language steps into formal logic and verifies them using a symbolic solver (Prover9). This pipeline rigorously verifies both local step validity and global proof reachability, achieving high agreement with human experts (98.80\% Step Accuracy and 95.22\% Proof Accuracy), thus ensuring a reliable metric for open-ended reasoning.

\textbf{Cognitive Assessment \& Insights.} 
Moving beyond aggregate accuracy metrics, we conduct a dual-faceted analysis of \textit{Convergent} and \textit{Divergent} thinking capabilities. We evaluate a diverse suite of state-of-the-art LLMs on LogicGraph, encompassing both general-purpose models, such as GPT-5.1~\cite{openai_gpt51}, and reasoning-oriented models, such as Gemini-3-Pro~\cite{google_gemini3_pro_preview}. Our results reveal a significant performance gap: while current models demonstrate competence in \textit{Convergent} metrics, their \textit{Divergent} thinking capabilities remain significantly constrained, characterized by a sharp decline in solution coverage as logical complexity increases. Although reasoning-oriented models generally outperform general models, the best model fails to achieve reliable exploratory reasoning. Furthermore, a fine-grained error analysis indicates that failures are predominantly ``result-oriented''; models frequently hallucinate intermediate lemmas to artificially force connections to target conclusions. These insights underscore the substantial gap between current model capabilities and the requirements for divergent thinking.

\section{Related Work}
\subsection{Logical Reasoning Datasets for LLMs} Logical reasoning evaluation for LLMs has moved from exam-style comprehension toward benchmarks that scale difficulty and require longer multi-step deductions. Early datasets such as LogiQA~\cite{liu2020logiqa}, AR-LSAT~\cite{zhong2021ar}, and ReClor~\cite{yu2020reclor} are derived from standardized tests. To better control reasoning depth and reduce linguistic confounds, synthetic benchmarks like RuleTaker~\cite{clark2020transformers} and ProofWriter~\cite{tafjord2021proofwriter} were introduced, and more recent resources provide explicit logical forms or harder multi-step settings~\cite{tian2021diagnosing, saparov2022language, morishita2023learning, han2024folio, parmar2024logicbench, chen2025justlogic}. However, most benchmarks score only the final answer, leaving the multi-path nature of logical derivation underexplored.

\begin{figure*}[t]
\centering
\includegraphics[width=\textwidth]{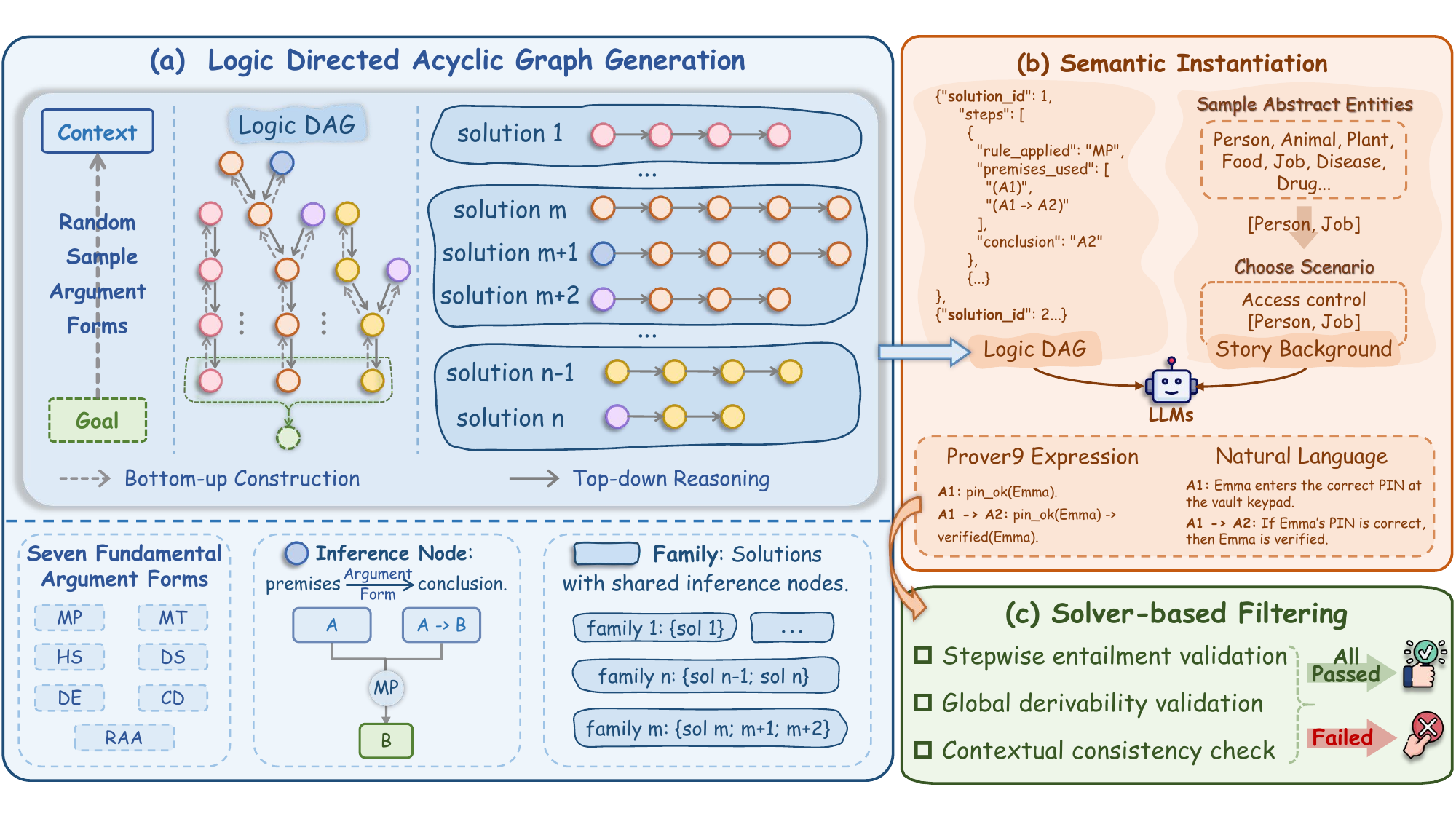}
\vspace{-25pt}
\caption{LogicGraph generation pipeline.
(a) Logic DAG generation (Section~\ref{sec:LogicDAG}) builds a goal-to-premise directed acyclic graph by sampling argument forms, yielding multiple valid reasoning paths; paths that share intermediate inference nodes are grouped into \emph{families}. 
(b) Semantic instantiation (Section~\ref{sec:semantic}) translates the DAG into Prover9 formulas and renders the corresponding steps into natural language by instantiating abstract entities and scenario context with LLMs. 
(c) Solver-based Filtering (Section~\ref{sec:quality}) uses Prover9 to validate each instance.}
\label{fig:generation}
\vspace{-5px}
\end{figure*}

\subsection{Symbolic Prover Augmented LLMs}
Symbolic provers (e.g., Z3~\cite{de2008z3} and proof assistants such as Lean~\cite{de2015lean}) offer trusted verification of formal statements, motivating neuro-symbolic pipelines that couple LLMs with prover checking.
Prior work typically uses LLMs to translate natural language into executable logical forms and delegate solving/verification to external engines~\cite{pan2023logic, deng2024enhancing, shen2025real, xu2024symbol, wang2024theoremllama}. Provers are exploited for data generation and filtering to ensure logical consistency of synthesized corpora~\cite{qi2025large}. 

\section{Task Formulation}
\label{sec:task_definition}
We formulate the reasoning task as finding all valid justifications for a goal hypothesis $\mathcal{G}$ given a set of premises $\mathcal{P}={p_1,p_2,\dots,p_n}$. We refer to each distinct justification as a \textit{derivation path}.

Formally, a derivation path is defined as a \textit{minimal support} set. A set $S\subseteq\mathcal{P}$ is a minimal support for $\mathcal{G}$, denoted by $\MinSup(S,\mathcal{G})$, if and only if (i) $S$ entails $\mathcal{G}$, (ii) no proper subset of $S$ entails $\mathcal{G}$:
\begin{equation}
S \vdash \mathcal{G}
\quad\text{and}\quad
\forall S' \subsetneq S:\ S' \nvdash \mathcal{G}.
\end{equation}

The task is to enumerate all such sets $S$ that satisfy $\MinSup(S,\mathcal{G})$. As shown in Figure~\ref{fig:example}, $S_{\text{escort}}=\{P3,P7\}$ is a minimal support for $\mathcal{G}$: removing either $P3$ or $P7$ breaks entailment. Other minimal supports correspond to alternative routes (e.g., PIN- or fingerprint-based).

\section{Automatic Dataset Generation Pipeline}
\label{sec:Generation Pipeline}
To ensure both logical rigor and linguistic diversity, we design a three-stage automated pipeline. Stage 1 constructs a symbolic directed acyclic graph to serve as a verifiable reasoning skeleton; Stage 2 translates this symbolic structure into coherent natural language contexts to simulate real-world complexity; Stage 3 employs a symbolic solver to verify the validity and consistency of the generated data, filtering out invalid samples.

\subsection{Symbolic Logic DAG Generation}
\label{sec:LogicDAG}
To mitigate logical inconsistencies in directly generated reasoning texts, we first construct a symbolic directed acyclic graph (Logic DAG) that specifies the ground-truth reasoning structure independent of natural language. As shown in Figure~\ref{fig:generation}(a), we build this graph via backward construction from a target conclusion, which will later be used for semantic instantiation.

\paragraph{Graph Primitives: Inference Node \& Family.}
A reasoning path is a sequence of \emph{inference nodes}, each being a rule instance $(\Gamma \Rightarrow \phi)$ that derives conclusion $\phi$ from a \emph{local} premise set $\Gamma$ (here $\Gamma$ is distinct from the global given premise pool $\mathcal{P}$ in Section~\ref{sec:task_definition}). We further group paths into \emph{families} if they share the same set of inference nodes.

\paragraph{Bottom-Up Construction.}
We begin by sampling a conclusion node. Then, for each current node $N$, we (i) sample a form from a predefined set of fundamental forms~\cite{Johnson1999-JOHALB-3} (e.g., Modus Ponens; formal definitions in Appendix~\ref{app:argument}), and (ii) generate its parent premises. For example, to derive $Q$ via Modus Ponens, we create node ($P \to Q$) and node $P$, and recursively expand the newly created premise nodes.

\paragraph{Multi-path Generation.}
To generate multiple reasoning paths, we first build a reasoning chain to a sampled depth, then pick an intermediate conclusion on the chain and expand it upward again using the same bottom-up procedure. Repeating this step produces a Logic DAG in which the goal is supported by multiple distinct premise sets. To ensure the ground-truth solution set is exhaustive, we assign a fresh atomic identifier to each newly introduced premise unless it is explicitly shared through an existing node, thereby avoiding unintended premise sharing and implicit extra paths.


\subsection{Semantic Instantiation}
\label{sec:semantic}
Although the Logic DAG ensures validity, it is not semantically rich enough for LLM evaluation. We thus transform each Logic DAG into a natural-language test case through an intermediate Prover9 form, which also enables downstream verification.
Following prior work, we define 32 abstract entity types (e.g., \textit{Person}, \textit{Job})~\cite{rosch1975family, wang2024can}. For each DAG, we sample a domain background, use Deepseek-V3.2-Exp to instantiate abstract symbols into domain-specific Prover9 predicates (e.g., convert A1 to $\text{pin\_ok(Emma)}$), and then verbalize the instantiated formulas into a coherent narrative while preserving all logical relations and avoiding unsupported facts.

\begin{figure*}[t]
\centering
\includegraphics[width=\textwidth]{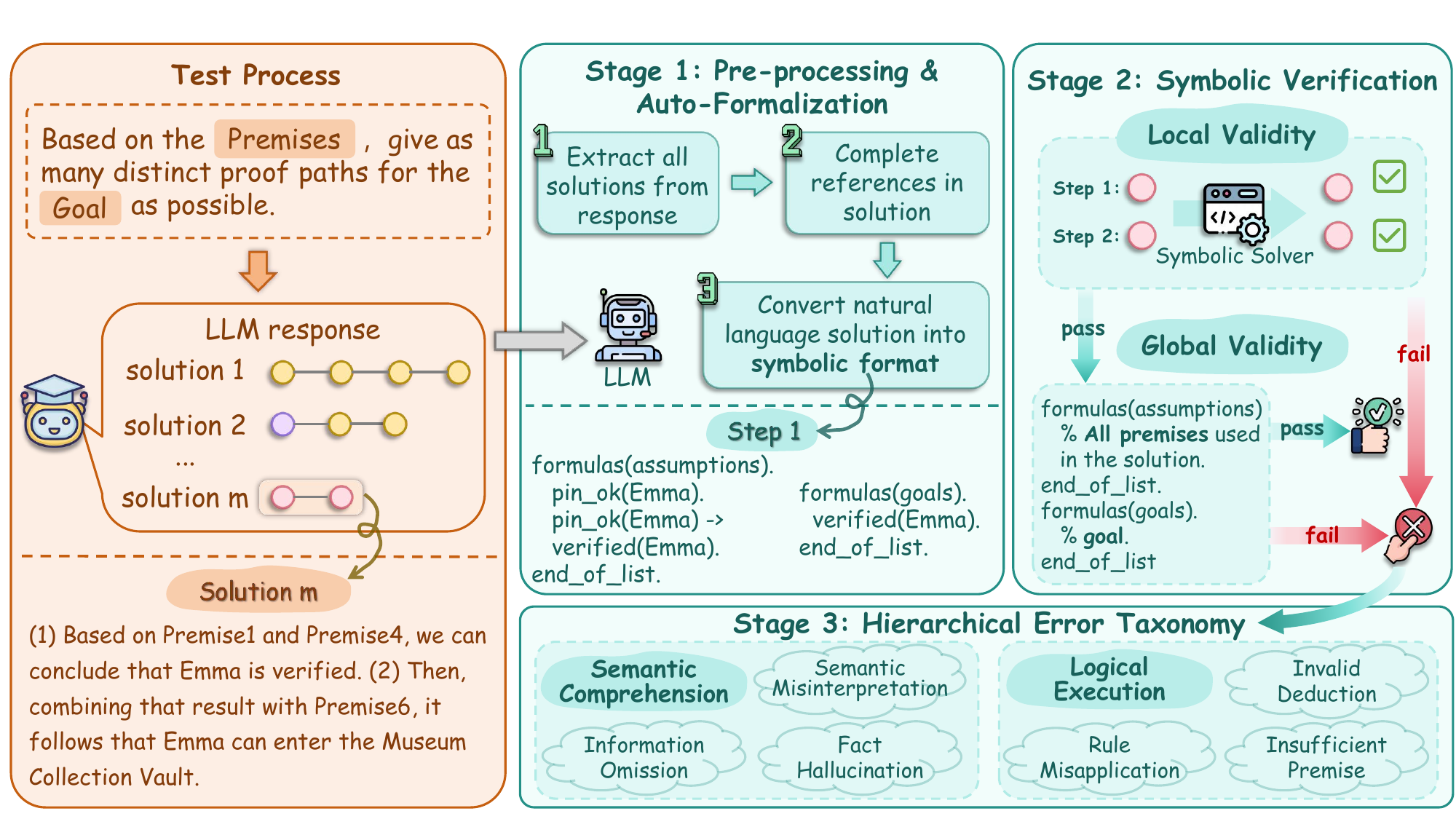}
\caption{Three-stage evaluation pipeline for LLM-generated multi-path proofs: (1) pre-processing and auto-formalization extract candidate solutions, resolve references, and translate natural-language steps into a symbolic representation (Prover9-style); (2) a solver is then used to validate local (stepwise) and global validity; (3) failures are annotated along two independent, non-exclusive error axes: semantic comprehension and logical execution.}
\label{fig:evaluation}
\vspace{-5pt}
\end{figure*}

\subsection{Solver-based Filtering}
\label{sec:quality}
Multi-path reasoning requires verification to ensure both edge-level (stepwise) soundness and graph-level consistency. Based on the Prover9 expressions produced in Sec.~\ref{sec:semantic}, we validate every sample with Prover9 and discard any sample that fails one of the following checks: (i) Stepwise Entailment. For each inference step (i.e., each edge in the Logic DAG), Prover9 verifies that the corresponding premises entail the derived statement, ensuring that every deduction is logically valid. (ii) Global Derivability. Prover9 confirms that all intermediate statements are derivable from the given premises and earlier derivations, and that the target goal is derivable from the full set of derivations, ensuring the graph is globally connected. (iii) Contextual Consistency. The union of all premises across multiple paths must be satisfiable (e.g., avoiding simultaneous $Fact(X)$ and $\neg Fact(X)$), ensuring no contradictions exist within the global context.




\subsection{Dataset Characteristics}
Based on the automatic generation pipeline described above, LogicGraph is scalable and robust against contamination. Using this dynamic pipeline, we curated a statistically robust benchmark of 900 instances via stratified sampling. We further divide the dataset into three difficulty tiers according to the number of valid derivation paths ($n$): \textit{Small} ($2 \le n \le 4$), \textit{Medium} ($5 \le n \le 7$), and \textit{Large} ($n \ge 8$), with 300 instances per tier.

As illustrated in Table~\ref{tab:comparison}, LogicGraph exhibits three structural properties \emph{induced by graph-based generation} that distinguish it from prior reasoning benchmarks: \textbf{(1) Multi-path High-depth.}
Unlike predominantly shallow and single-path datasets, each LogicGraph query admits 2--19 \emph{valid} proof paths with an average depth of 6.01. This requires not only long-chain deduction but also exploration over multiple proof trajectories.
\textbf{(2) Inference Node Reuse.}
Unlike chain-style proofs where intermediate conclusions are path-specific, LogicGraph allows intermediate inference nodes to be shared across branches (Reuse Ratio: 1.0--1.9). Consequently, models must identify and reuse shared subproofs rather than solving each path independently.
\textbf{(3) Inherent Logical Distractions.}
Beyond static semantic noise or dead-end facts, LogicGraph introduces structural distractions from the DAG itself: a premise can be crucial for one valid path yet distracting for another. Models must selectively activate the premises supporting a target trajectory from logically valid but competing evidence.

\section{Neuro-Symbolic Evaluation Framework}
\label{sec:Evaluation Pipeline}
The rapid advancement of LLMs in complex reasoning necessitates a more rigorous evaluation paradigm. Current evaluation methods often rely on LLM-as-a-Judge. Yet this purely neural paradigm inherits the same weaknesses it aims to diagnose—hallucinations, weak formal grounding, and systematic biases in self-correction—making it easy to miss subtle logical fallacies \cite{szymanski2025limitations, wang2025trustjudge, feng2025we}. Conversely, traditional symbolic solvers provide deterministic verification but remain too rigid to handle the informal reasoning traces.

To overcome these limitations, we integrate an LLM as a flexible formal translator and a symbolic solver as a rigorous verifier~\cite{bhuyan2024neuro}. In the following sections, we detail the evaluation pipeline, define the metrics, and present a meta-evaluation to validate the framework's reliability.

\subsection{The Evaluation Pipeline}
\label{sec:eva_method}
As illustrated in Figure~\ref{fig:evaluation}, our framework evaluates the logical validity of the entire reasoning process via a three-stage pipeline.
\paragraph{Stage 1: Pre-processing \& Auto-Formalization.}
To support symbolic verification, we disentangle distinct derivation paths from the model output and reconstruct implicit dependencies (e.g., ``From the previous step…'') into an explicit chain. 
An LLM then translates each NL step into Prover9 syntax. To ensure fidelity, we use in-context examples aligning Prover9 expressions from the data synthesis stage with their corresponding NL descriptions.

\paragraph{Stage 2: Symbolic Verification.}
We employ Prover9 to verify the reasoning chain at two levels:
(i) \textit{Local Validity} checks whether each step $S_t$ logically follows from its cited premises $P_t$ (i.e., $P_t \vdash S_t$), (ii) \textit{Global Validity} ensures whether the final conclusion $G$ can be derived solely from the subset of premises explicitly used in the solution. Together, these checks ensure grounded proofs and flag hallucinated steps.

\paragraph{Stage 3: Hierarchical Error Taxonomy.}
To pinpoint where and why models fail, we propose a two-dimensional error taxonomy (Table~\ref{tab:error_taxonomy_main}):

\noindent \textit{Semantic Comprehension.} Faithfulness to the provided context, for example,  when the model hallucinates a relationship not stated in the premises.

\noindent \textit{Logical Execution.} Validity of derivation steps, such as applying an incorrect rule or producing a non-sequitur conclusion.

\begin{table*}[]
\centering
\begin{adjustbox}{width=\textwidth}
\renewcommand{\arraystretch}{1.1}
\begin{tabular}{cll}
\hline
\rowcolor{gray!10}
\rowcolor[HTML]{EFEFEF} 
\textbf{Dimension} & \textbf{Error Type} & \multicolumn{1}{c}{\cellcolor[HTML]{EFEFEF}\textbf{Definition}} \\ \hline
 & Semantic Misinterpretation & \textit{Incorrectly understood the meaning of facts or rules.} \\
 & Information Omission & \textit{Ignored necessary premises or constraints explicitly stated in the context.} \\
\multirow{-3}{*}{\textbf{\begin{tabular}[c]{@{}c@{}}Semantic\\ Comprehension\end{tabular}}} & Fact Hallucination & \textit{Used external or fabricated premises not present in the input.} \\
\hline
 & Invalid Deduction & \textit{The conclusion does not logically follow from the cited premises.} \\
 & Rule Misapplication & \textit{Applied a rule that doesn't match the current premises.} \\
\multirow{-3}{*}{\textbf{\begin{tabular}[c]{@{}c@{}}Logical\\ Execution\end{tabular}}}
& Insufficient Premise & \textit{Cited relevant premises but incomplete for deriving the conclusion.} \\
\hline
\end{tabular}
\end{adjustbox}
\label{tab:error_taxonomy_main}
\vspace{-5pt}
\caption{The proposed hierarchical error taxonomy. The taxonomy distinguishes between failures in processing input information (Comprehension) and failures in the reasoning process (Execution).}
\label{tab:error_taxonomy_main}
\vspace{-5pt}
\end{table*}

\subsection{Cognitive Evaluation Metrics}
\label{metric}
To capture the multifaceted nature of LLM reasoning, we adopt a two-axis evaluation protocol inspired by Guilford's Structure of Intellect~\cite{guilford1967nature}, evaluating convergent and divergent thinking beyond outcome accuracy.

Convergent Thinking assesses the model's ability to derive a single correct conclusion through valid logical steps. We quantify this via three primary dimensions: (i) \textit{Success Rate}, the proportion of test cases with at least one valid proof path; (ii) \textit{Precision}, the ratio of valid solutions to total generations, reflecting hallucination resistance; and (iii) \textit{Shortest Path Finding Rate}, the percentage of solutions matching the minimum ground-truth step count, which measures reasoning conciseness.

Divergent Thinking evaluates creativity and flexibility in discovering multiple distinct paths. This is measured by: (i) \textit{Diversity (Solution Recall)}, defined as $R_{\text{sol}} = |\mathcal{S}_{Model} \cap \mathcal{S}_{GT}| / |\mathcal{S}_{GT}|$, which quantifies the coverage of the solution space; (ii) \textit{Versatility (Family Recall)}, which reflects the agility to switch between distinct reasoning strategies; and (iii) \textit{Originality}, which highlights the ability to identify rare paths by calculating the inverse frequency of a solution's discovery across all models.

\begin{table}[h]
\label{tab:meta_eval}
\begin{adjustbox}{width=\columnwidth}
\centering
\begin{tabular}{lccc}
\hline
\textbf{Evaluator} & \textbf{Reference} & \textbf{Acc(S)} & \textbf{Acc(P)} \\
\hline
\multicolumn{4}{c}{LLM-as-a-Judge Baselines}
\\
\hline
Deepseek-V3.2-Exp & Required & 77.25 & 71.46 \\
Deepseek-V3.2-T$^{*}$ & Required & 87.19  & 83.59\\
Gemini-3-Pro & Required & 86.11 & 83.91 \\
\hline
\multicolumn{4}{c}{Neuro-Symbolic} \\
\hline
\textbf{Ours(Gemini)} & \textbf{None} & \textbf{97.57} & \textbf{94.85} \\
\textbf{Ours(Deepseek)} & \textbf{None} & \textbf{98.80} & \textbf{95.22} \\
\hline
\end{tabular}
\end{adjustbox}
\vspace{-5pt}
\caption{Meta-evaluation results comparing agreement with human annotators. Acc(S) and Acc(P) denote per-step and overall proof validity, respectively.}
\label{tab:meta_eval}
\vspace{-5px}
\end{table}

\subsection{Evaluator Reliability Validation}
To validate our framework, we measure agreement with expert human judgments on proof validity and compare against reference-based LLM-as-a-Judge baselines. As shown in Table~\ref{tab:meta_eval}, our reference-free neuro-symbolic evaluator (with Prover9) is robust across backbone LLMs: both the Gemini-2.5-Flash and DeepSeek-V3.2-Exp versions outperform reference-based LLM-as-a-Judge baselines, with DeepSeek-V3.2-Exp achieving the highest agreement. This suggests symbolic verification reduces false positives from fluent-but-invalid traces.

\newcommand{\celltriple}[4]{\makecell{\textbf{#1}\\{\tiny #2/#3/#4}}}
\begin{table*}[t]
\centering
\begin{adjustbox}{width=\textwidth}
\scriptsize
\setlength{\tabcolsep}{3pt}
\renewcommand{\arraystretch}{1.25}
\begin{tabular}{p{2.4cm} | ccc| ccc| c}
\toprule
\multirow{2}{*}{\textbf{Model}} &
\multicolumn{3}{c|}{\textbf{Convergent} ($\uparrow$)} &
\multicolumn{3}{c|}{\textbf{Divergent} ($\uparrow$)} &
\multirow{2}{*}{\textbf{\makecell{Token Eff.\\(Overall)$\downarrow$}}} \\
& \makecell{\textbf{SR}\\\tiny Avg; S/M/L}
& \makecell{\textbf{Prec.}\\\tiny Avg; S/M/L}
& \makecell{\textbf{SPF}\\\tiny Avg; S/M/L}
& \makecell{\textbf{Div.}\\\tiny Avg; S/M/L}
& \makecell{\textbf{Vers.}\\\tiny Avg; S/M/L}
& \makecell{\textbf{Orig.}\\\tiny Avg; S/M/L}
& \\
\midrule
\rowcolor{gray!10}\multicolumn{8}{c}{\textbf{General Models}}\\
\midrule
GLM-4.6
& \celltriple{21.11}{19.67}{21.00}{22.67}
& \celltriple{8.47}{11.20}{7.62}{6.58}
& \celltriple{14.00}{15.00}{13.00}{14.00}
& \celltriple{6.19}{9.83}{5.30}{3.45}
& \celltriple{9.66}{13.50}{8.35}{7.14}
& \celltriple{2.64}{1.85}{2.71}{3.36}
& 7274.96 \\
GPT-OSS-120B
& \celltriple{12.22}{13.67}{11.33}{11.67}
& \celltriple{9.97}{12.07}{8.06}{9.79}
& \celltriple{7.89}{9.67}{8.67}{5.33}
& \celltriple{3.09}{5.75}{2.12}{1.41}
& \celltriple{5.50}{9.75}{3.76}{2.99}
& \celltriple{1.21}{1.48}{1.06}{1.08}
& 27120.45 \\
GPT-5.1
& \celltriple{21.56}{19.67}{20.67}{24.33}
& \celltriple{9.69}{10.94}{8.76}{9.37}
& \celltriple{13.56}{14.00}{13.67}{13.00}
& \celltriple{5.98}{8.86}{5.02}{4.07}
& \celltriple{9.96}{13.11}{8.39}{8.39}
& \celltriple{3.20}{1.94}{3.03}{4.63}
& 7096.54 \\
Qwen3-235B-A22B
& \celltriple{29.56}{30.67}{27.67}{30.33}
& \celltriple{13.89}{17.47}{12.69}{11.52}
& \celltriple{17.33}{20.67}{17.00}{14.33}
& \celltriple{8.23}{13.28}{6.60}{4.80}
& \celltriple{14.65}{21.67}{12.14}{10.14}
& \celltriple{4.02}{3.49}{3.46}{5.11}
& 7401.54 \\
Claude-Sonnet-4.5
& \celltriple{53.11}{52.00}{51.67}{55.67}
& \celltriple{36.05}{39.60}{34.09}{34.47}
& \celltriple{29.22}{31.33}{30.67}{25.67}
& \celltriple{12.88}{19.94}{10.97}{7.74}
& \celltriple{25.42}{37.00}{21.96}{17.31}
& \celltriple{8.26}{7.24}{7.83}{9.7}
& 3431.10 \\
\textbf{Deepseek-V3.2-Exp}
& \celltriple{61.00}{57.33}{61.67}{64.00}
& \celltriple{52.04}{49.34}{49.90}{56.89}
& \celltriple{34.44}{33.67}{35.33}{34.33}
& \celltriple{16.54}{25.39}{14.51}{9.71}
& \celltriple{31.38}{43.67}{28.64}{21.84}
& \celltriple{10.20}{8.48}{10.32}{11.79}
& 3397.51 \\
\midrule
\rowcolor{gray!10}\multicolumn{8}{c}{\textbf{Reasoning-Oriented Models}}\\
\midrule
o3
& \celltriple{58.33}{54.33}{55.00}{65.67}
& \celltriple{50.79}{47.77}{47.39}{57.22}
& \celltriple{39.44}{43.00}{38.33}{37.00}
& \celltriple{17.94}{28.28}{14.70}{10.84}
& \celltriple{30.02}{41.33}{25.71}{23.01}
& \celltriple{9.31}{7.51}{8.50}{11.991}
& 9446.91 \\
o4-mini
& \celltriple{57.78}{56.67}{57.33}{59.33}
& \celltriple{51.38}{48.99}{49.40}{55.76}
& \celltriple{33.67}{38.00}{32.33}{30.67}
& \celltriple{16.17}{25.14}{14.04}{9.34}
& \celltriple{29.20}{40.42}{27.21}{19.97}
& \celltriple{8.64}{7.77}{8.84}{9.31}
& 6903.90 \\
QwQ-32B
& \celltriple{50.89}{48.00}{49.00}{55.67}
& \celltriple{25.91}{27.66}{24.25}{25.81}
& \celltriple{35.44}{34.33}{37.33}{34.67}
& \celltriple{16.20}{23.81}{14.48}{10.32}
& \celltriple{27.38}{34.97}{25.13}{22.04}
& \celltriple{7.64}{6.06}{7.53}{9.32}
& 12228.78 \\
Qwen3-235B-A22B-T$^{*}$
& \celltriple{91.22}{85.00}{93.33}{95.33}
& \celltriple{76.89}{73.08}{78.49}{79.11}
& \celltriple{70.22}{68.00}{75.33}{67.33}
& \celltriple{45.83}{59.42}{45.42}{32.64}
& \celltriple{69.12}{76.67}{69.28}{61.42}
& \celltriple{21.61}{16.05}{21.98}{26.79}
& 5940.66 \\
Kimi-K2-T$^{*}$
& \celltriple{72.22}{69.67}{74.67}{72.33}
& \celltriple{48.20}{46.51}{50.67}{47.42}
& \celltriple{51.44}{56.00}{54.33}{44.00}
& \celltriple{29.04}{42.06}{27.68}{17.37}
& \celltriple{44.94}{58.08}{44.52}{32.22}
& \celltriple{13.29}{10.71}{14.40}{14.77}
& 13764.23 \\
Claude-Sonnet-4.5-T$^{*}$
& \celltriple{61.67}{61.67}{59.67}{63.67}
& \celltriple{47.82}{50.58}{46.05}{46.82}
& \celltriple{34.22}{40.00}{34.00}{28.67}
& \celltriple{15.66}{25.47}{13.07}{8.43}
& \celltriple{30.15}{45.31}{26.23}{18.92}
& \celltriple{9.72}{8.63}{9.43}{11.1}
& 3476.57 \\
Deepseek-V3.2-Exp-T$^{*}$
& \celltriple{88.00}{82.00}{89.33}{92.67}
& \celltriple{76.96}{71.02}{77.83}{82.04}
& \celltriple{66.33}{67.33}{69.67}{62.00}
& \celltriple{39.76}{53.81}{39.22}{26.25}
& \celltriple{61.02}{70.67}{61.32}{51.07}
& \celltriple{18.72}{14.47}{18.97}{22.71}
& 5582.40 \\
Gemini-2.5-Flash
& \celltriple{85.56}{82.00}{85.00}{89.67}
& \celltriple{67.04}{63.82}{67.16}{70.14}
& \celltriple{65.89}{67.00}{66.33}{64.33}
& \celltriple{41.35}{53.28}{40.43}{30.33}
& \celltriple{61.43}{69.61}{60.44}{54.24}
& \celltriple{20.34}{14.97}{19.49}{26.56}
& 6835.60 \\
Gemini-2.5-Pro
& \celltriple{87.78}{82.67}{88.67}{92.00}
& \celltriple{73.38}{72.20}{72.90}{75.03}
& \celltriple{72.78}{72.67}{75.67}{70.00}
& \celltriple{51.07}{60.56}{53.11}{39.53}
& \celltriple{69.13}{74.50}{69.83}{63.07}
& \celltriple{23.77}{16.17}{24.33}{30.81}
& 4414.16 \\
\textbf{Gemini-3-Pro-Preview}
& \celltriple{96.11}{93.00}{98.00}{97.33}
& \celltriple{90.10}{89.11}{91.89}{89.31}
& \celltriple{80.33}{80.00}{82.67}{78.33}
& \celltriple{59.60}{70.03}{60.30}{48.47}
& \celltriple{79.44}{83.03}{79.99}{75.30}
& \celltriple{29.31}{20.96}{30.3}{36.67}
& 1302.15 \\
\bottomrule
\end{tabular}
\end{adjustbox}
\caption{Comprehensive model comparison on the LogicGraph, covering convergent metrics (Success Rate/Precision/Short Path Finding Rate), divergent metrics (Diversity/Versatility/Originality), and overall Token Efficiency ($\downarrow$) across Small/Medium/Large. In each cell, \textbf{Avg} (over Small/Medium/Large) is bolded; the second line reports Small/Medium/Large in order. -T$^{*}$ denotes Thinking.}
\label{tab:main_re}
\vspace{-5px}
\end{table*}

\section{Experiments}
\subsection{Experimental Setting}
We conduct comprehensive experiments on LogicGraph to evaluate model performance across varying degrees of logical complexity.

\paragraph{Baselines.}
We evaluate several SOTA LLMs on LogicGraph, including both proprietary APIs and open-weight releases, including GLM-4.6~\cite{zhipu_glm46}, GPT-5.1~\cite{openai_gpt51}, Claude-Sonnet-4.5~\cite{anthropic_claude}, o3/o4-mini~\cite{openai_o3}, Gemini 2.5/3~\cite{google_gemini25_flash,google_gemini25_pro,google_gemini3_pro_preview}, as well as open source models such as GPT-OSS-120B~\cite{openai2025gptoss120bgptoss20bmodel}, Qwen3-235B-A22B (and its Thinking variant)~\cite{yang2025qwen3}, QwQ-32B~\cite{qwq_32b}, and DeepSeek-V3.2-Exp (and its Thinking variant)~\cite{deepseek_v32exp}. For all models, we use a unified prompting protocol that encourages the model to produce as many independent and verifiable solution paths as possible under a fixed structured answer template; prompt templates are given in Appendix~\ref{app:prompts}.

\paragraph{Evaluation Flow.}
Following the evaluation results in Sec.~\ref{sec:eva_method}, and considering the accuracy--cost trade-off in deployment, we select DeepSeek-V3.2-Exp as the default backbone model for our experiments; see Appendix~\ref{app:evaluator} for the detailed prompt.

\subsection{Main Results}
Table~\ref{tab:main_re} presents the comparative analysis of General and Reasoning-Oriented models on LogicGraph. We synthesize the experimental results into three key observations:

\paragraph{Reasoning-oriented models achieve stronger performance without extra cost.}
Overall, reasoning-oriented models outperform general models on convergent metrics, indicating that stronger inference is essential for logic tasks. Notably, Gemini-3-Pro-Preview establishes a new state-of-the-art on all metrics, while also exhibiting the highest token efficiency, suggesting that improved reasoning reduces wasted generation on incorrect branches and reaches valid reasoning paths more directly.

\paragraph{Top models remain concentrated on a narrow subset of valid reasoning paths.}
Despite near-saturated Success Rates for top models, divergent metrics remain markedly lower, revealing a gap between finding one valid reasoning path and enumerating many. This indicates that current models tend to concentrate on a limited subset of high-probability paths rather than systematically exploring diverse alternatives.

\begin{figure*}[t]
\centering
\includegraphics[width=\textwidth]{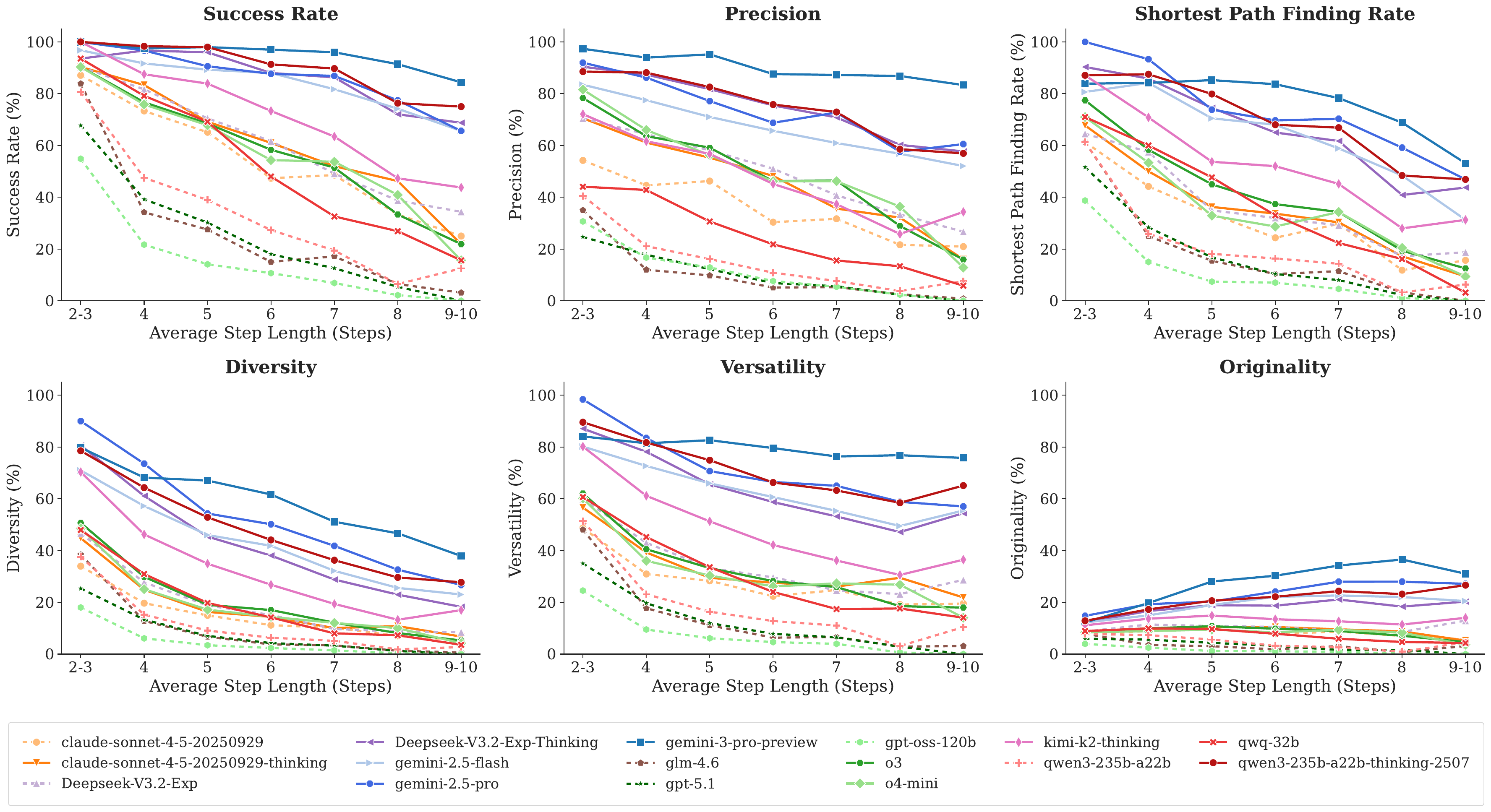}
\vspace{-15pt}
\caption{Performance dynamics of LLMs across varying reasoning depths. Solid lines represent Reasoning-oriented LLMs; dashed lines represent General-purpose LLMs.}
\label{fig:step}
\vspace{-5px}
\end{figure*}

\begin{figure}[t]
\centering
\includegraphics[width=\columnwidth]{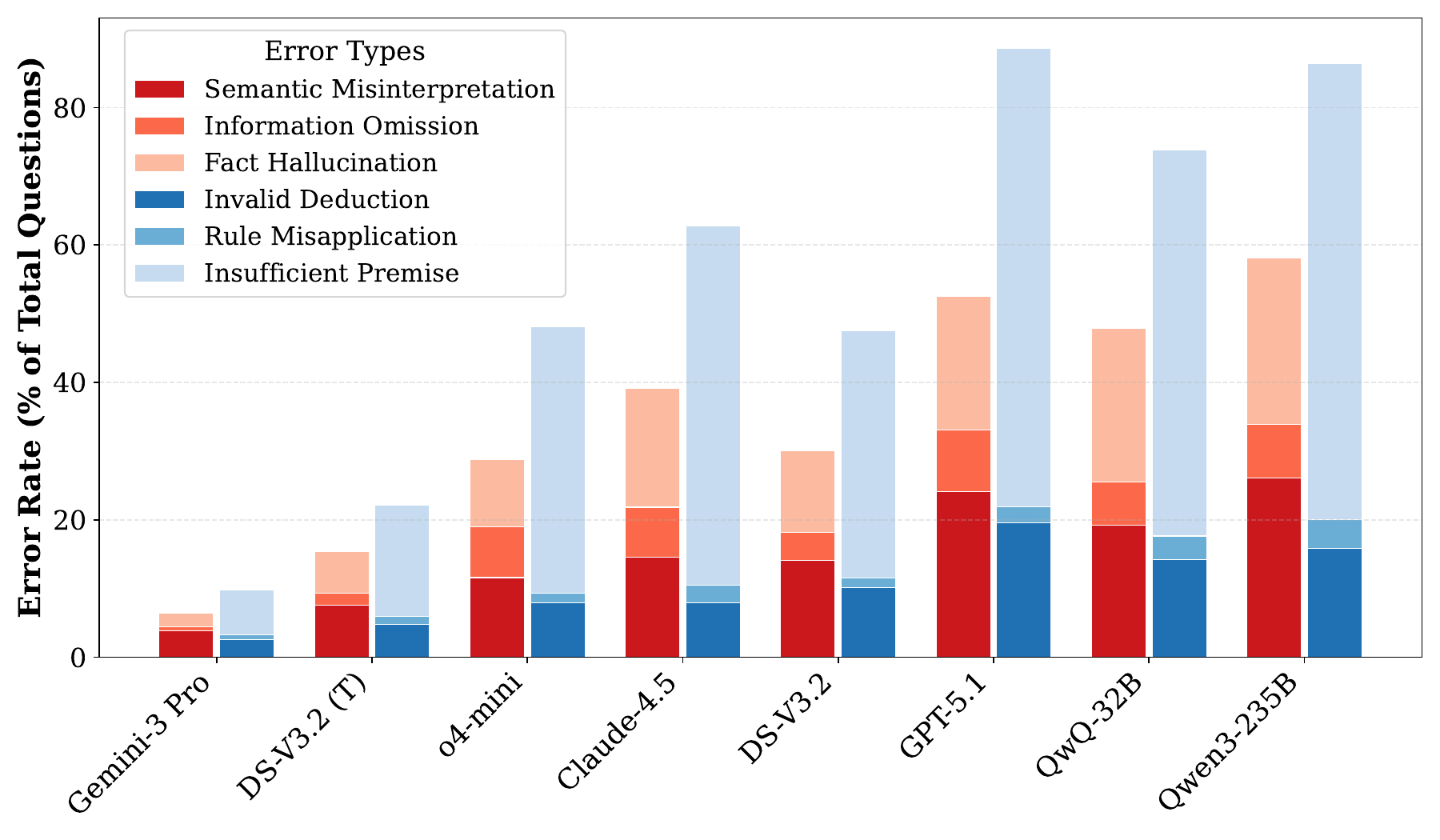}
\vspace{-25pt}
\caption{Comparative Analysis of Error Type: Semantic Comprehension vs. Logical Execution.}
\label{fig:eror_dis_main}
\vspace{-5px}
\end{figure}

\paragraph{Multiplicity: Models can switch strategies, but cannot enumerate many valid paths.}
As the number of valid reasoning paths increases from Small to Large, both Versatility and Diversity decline across models. However, Versatility is typically more robust than Diversity, indicating that models can still reach multiple high-level strategies but increasingly fail to exhaustively list the many concrete variants implied by each strategy.

\subsection{Depth-Induced Performance Degradation}
As shown in figure \ref{fig:step}, as reasoning depth increases, both success rate and precision decline steadily, indicating that errors accumulate and amplify across multi-hop chains. Reasoning models remain more robust in the mid-to-high step ranges, while general models degrade more sharply, suggesting reasoning optimization strengthens cross-hop consistency and error tolerance. Beyond correctness, Diversity and Versatility often drop earlier and faster, implying that deeper puzzles drive models toward narrower reasoning patterns and explore fewer alternatives.

\subsection{Analysis of Error Categories}
Figure~\ref{fig:eror_dis_main} presents the error types distribution across representative models (Appendix~\ref{app:error_taxonomy} provides results for all evaluated models), while most models maintain relatively low error rates in semantic comprehension, they exhibit a pronounced spike in logical execution errors, particularly invalid deduction and insufficient premise. This suggests that in proof tasks, LLMs may generate result-oriented fabrications. When a valid logical chain from premises to conclusion is unavailable, they tend to invent intermediate lemmas or make unjustified leaps to produce a superficially complete proof.

\subsection{Case Study}
\begin{figure}[t]
\centering
\includegraphics[width=\columnwidth]{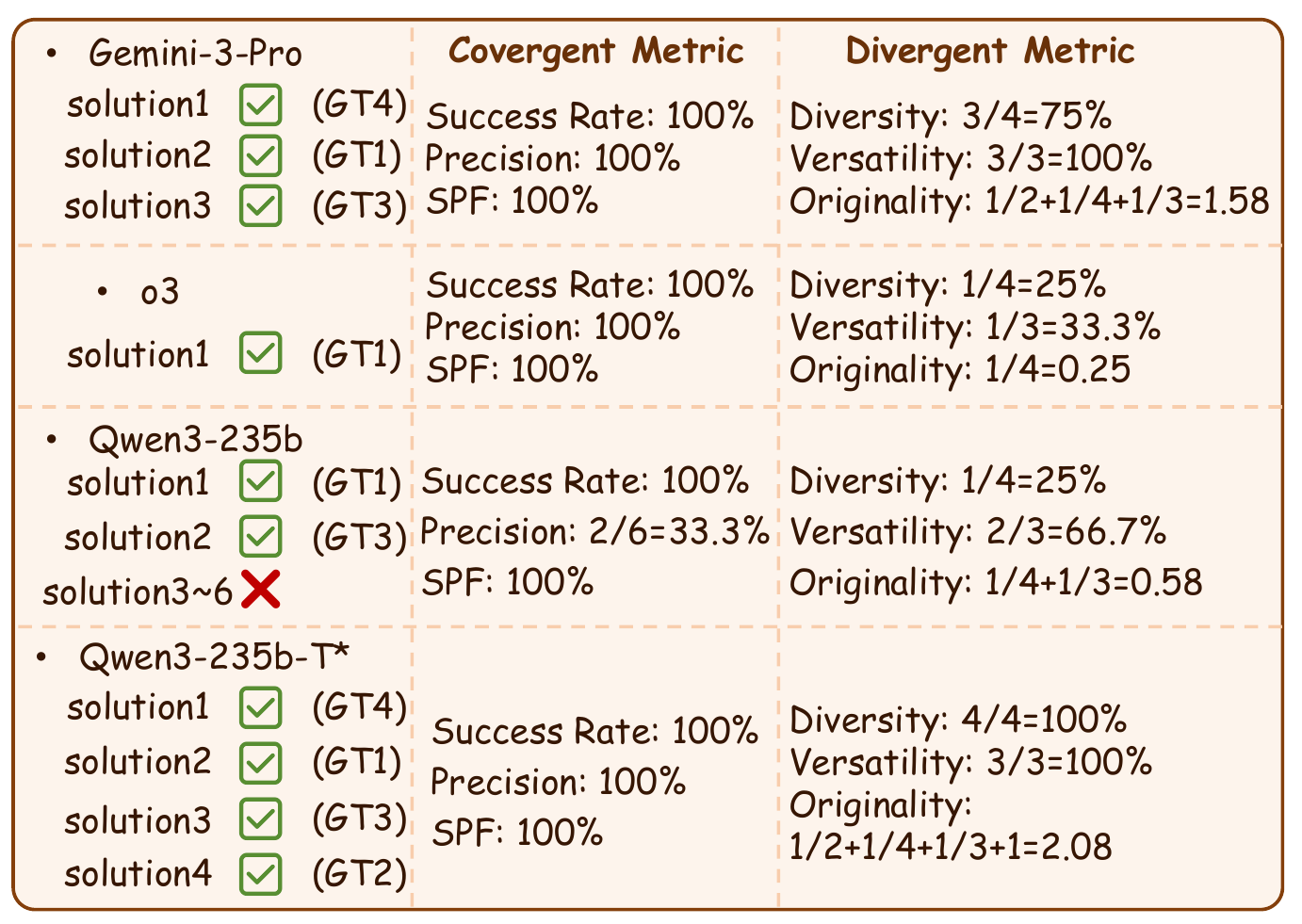}
\vspace{-15pt}
\caption{Model performance on case (Figure~\ref{fig:sexam}). GT represents the ground truth paths.}
\label{fig:case}
\vspace{-10px}
\end{figure}

Figure~\ref{fig:case} shows why we need Divergent Metrics beyond standard scores. Further analysis uncovers hidden reasoning flaws even when answers are correct; see Appendix~\ref{app:case1} for more details.

\section{Conclusion}

We develop a neuro-symbolic generation and solver-verification pipeline for constructing high-depth multi-path logical reasoning problems. Based on it, we build LogicGraph, the first benchmark where each instance comes with an exhaustive set of minimal proofs and logical distractions. Evaluations of state-of-the-art LLMs reveal a clear limitation: models often fixate on one route and increasingly miss alternatives as depth grows. LogicGraph offers a testbed to drive future models toward more flexible, comprehensive reasoning.

\section*{Limitations}
While LogicGraph represents a significant step towards evaluating complex reasoning, we acknowledge the following limitations:

\paragraph{Synthetic-to-Real Gap.} Although we employ advanced LLMs to render symbolic logic into natural language narratives, the dataset remains synthetically constructed. Real-world reasoning often involves ambiguity, fuzzy logic, and probabilistic inference, which are strictly controlled in our current discrete logic framework. The clean logical separation in our benchmark may not fully reflect the noisy information retrieval challenges found in the real world.

\paragraph{Computational Cost.} The multi-path problems neuro-symbolic verification process is computationally intensive compared to simple multiple-choice evaluations. This may pose challenges for rapid, low-resource model evaluation.

\section*{Ethical Statement}
LogicGraph is a fully synthetic benchmark built from abstract entities and explicit logic rules, so it contains no personally identifiable information or copyright-protected text, eliminating risks related to data privacy or intellectual property infringement. A remaining concern is that LLM-based surface realization can introduce latent social biases; we reduce this risk by using constrained templates and distributing instances across multiple neutral domains to avoid skewed coverage. We also acknowledge the environmental cost of multi-path reasoning and neuro-symbolic verification, which can be more computationally intensive than standard evaluations; future work will prioritize more efficient pipelines to reduce the associated carbon footprint. Although stronger reasoning may be repurposed in dual-use settings, the benchmark’s abstract, formal scope limits direct harmful applicability. We will release the LogicGraph benchmark artifacts, including dataset instances, generation scripts, and evaluation code, under a permissive open-source license to facilitate reuse and reproducibility.

\bibliography{main}
\newpage

\appendix
\section{Benchmark}
\subsection{Argument Forms}
\label{app:argument}
In this paper, we adopt the seven most basic argument forms~\cite{Johnson1999-JOHALB-3}, as summarized in Table~\ref{tab:basic-forms}. Using these core rules, many inference patterns that are not listed explicitly can often be obtained by composing several basic forms. As an illustration, we show how the Destructive Dilemma (DD) can be derived using only the argument forms in Table~\ref{tab:basic-forms}.

\noindent\textbf{Derived rule (DD).}\;
\[
\text{DD:}\quad ((p\to q)\land(r\to s)\land(\neg q \lor \neg s))\vdash(\neg p \lor \neg r).
\]

\noindent\textbf{Derivation using only the forms in Table~X.}\;
\[
\begin{array}{rll}
1. & p\to q & \text{Premise} \\
2. & r\to s & \text{Premise} \\
3. & \neg q \lor \neg s & \text{Premise} \\
4. & \neg q \to \neg p & \text{from 1 by MT} \\
5. & \neg s \to \neg r & \text{from 2 by MT} \\
6. & \neg p \lor \neg r & \text{from 3, 4, 5 by CD} \\
\end{array}
\]

\begin{table*}[t]
\begin{adjustbox}{width=\textwidth}
\renewcommand{\arraystretch}{1.25}
\centering
\begin{tabular}{lll}
\hline
\textbf{Argument Form} & \textbf{Formal Notation} & \textbf{Definition (informal)} \\
\hline
Modus Ponens (MP)
& $((p \to q)\land p)\vdash q$
& If $p\to q$ and $p$, then $q$. \\

Modus Tollens (MT)
& $((p \to q)\land \neg q)\vdash \neg p$
& If $p\to q$ and $\neg q$, then $\neg p$. \\

Hypothetical Syllogism (HS)
& $((p \to q)\land(q \to r))\vdash (p \to r)$
& If $p\to q$ and $q\to r$, then $p\to r$. \\

Disjunctive Syllogism (DS)
& $((p \lor q)\land \neg p)\vdash q$
& If $p\lor q$ and $\neg p$, then $q$ (symmetrically, $\neg q$ yields $p$). \\

Constructive Dilemma (CD)
& $((p \to q)\land(r \to s)\land(p \lor r))\vdash (q \lor s)$
& If $p\to q$, $r\to s$, and $p\lor r$, then $q\lor s$. \\

Reductio ad Absurdum (RAA)
& $((p \to q)\land(p \to \neg q))\vdash \neg p$
& If assuming $p$ leads to a contradiction (e.g., both $q$ and $\neg q$), then $\neg p$. \\

Disjunction Elimination (DE)
& $((p \lor q)\land(p \to r)\land(q \to r))\vdash r$
& If $p\lor q$, and $p\to r$ and $q\to r$, then $r$. \\
\hline
\end{tabular}
\end{adjustbox}
\caption{Basic argument forms and their formal notations.}
\label{tab:basic-forms}
\end{table*}

\subsection{Dataset Distribution}
\begin{figure}[t]
\centering
\includegraphics[width=0.9\linewidth]{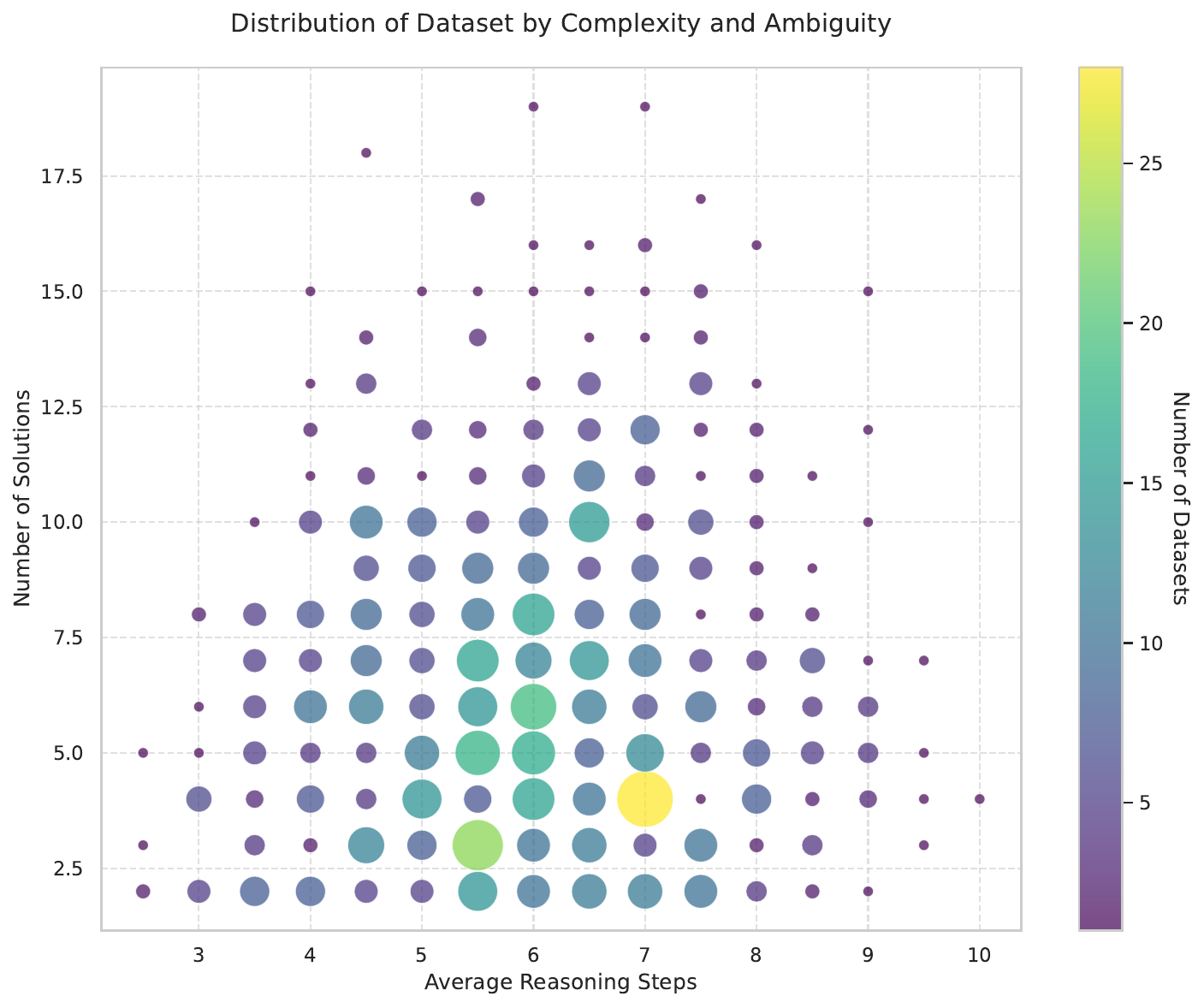}
\caption{Distribution of the benchmark dataset by average steps and number of solutions. The bubble size indicates the frequency of samples.}
\label{fig:data_dis}
\end{figure}

Figure~\ref{fig:data_dis} visualizes the LogicGraph's distribution along two key dimensions: (1) the average number of reasoning steps (x-axis), which reflects reasoning complexity, and (2) the number of valid ground-truth solutions (y-axis), which reflects solution-space ambiguity. The bubble size and color both indicate the sample frequency at each coordinate.

A notable observation is that the two dimensions are largely independent. The Pearson correlation between average reasoning steps and the number of valid solutions is approximately $\rho \approx 0.06$, indicating a negligible linear relationship. This suggests that our construction effectively decouples reasoning depth from the size of the solution space, enabling controlled evaluation of both complexity and ambiguity within a single benchmark.

\subsection{Dataset Domain Diversity Analysis}
As illustrated in Table~\ref{tab:domain_distribution}, LogicGraph encompasses over 90 distinct specific fields spanning 7 major categories: Business \& Finance, Environment \& Science, Medical \& Health, Law \& Governance, Engineering \& Tech, Humanities \& Society, and Transportation \& Urban. The distribution includes high-stakes environments such as Emergency Medicine, Clinical Trials, and Network Security, as well as complex professional scenarios like Regulatory Compliance, Industrial Engineering, and Legal Procedures. This diversity ensures that the evaluation reflects the model's ability to handle specialized knowledge and reasoning patterns inherent to different professional and practical domains.

\begin{table*}[t]
\centering
\label{tab:domain_distribution}
\resizebox{\textwidth}{!}{%
\begin{tabular}{l l r r}
\toprule
\textbf{Domain Category} & \textbf{Specific Fields (Examples)} & \textbf{\# Samples} & \textbf{Percentage} \\
\midrule
Business \& Finance & Quality Control, Insurance, Industrial Engineering, etc. & 207 & 23.0\% \\
Environment \& Science & Scientific Method, Biodiversity, Forestry, etc. & 187 & 20.8\% \\
Medical \& Health & Clinical Trials, Pharmacology, Emergency Medicine, etc. & 167 & 18.6\% \\
Law \& Governance & Regulatory Compliance, Legal Procedure, Litigation, etc. & 113 & 12.6\% \\
Engineering \& Tech & IoT, Software Development, Telecommunications, etc. & 94 & 10.4\% \\
Humanities \& Society & Sports, Visual Arts, Culinary Arts, etc. & 84 & 9.3\% \\
Other & General Knowledge, Mixed Domains & 25 & 2.8\% \\
Transportation \& Urban & Soil Science, Agriculture, Automotive, etc. & 23 & 2.6\% \\
\bottomrule
\end{tabular}
}
\caption{Domain Distribution of the LogicGraph Benchmark. The dataset covers a wide spectrum of real-world scenarios, categorized into 7 major domains and over 90 specific fields. This diversity ensures a comprehensive evaluation of reasoning capabilities across various professional and practical contexts.}
\label{tab:domain_distribution}
\end{table*}

\subsection{Examples}
\begin{figure*}[t]
\centering
\includegraphics[width=0.89\textwidth]{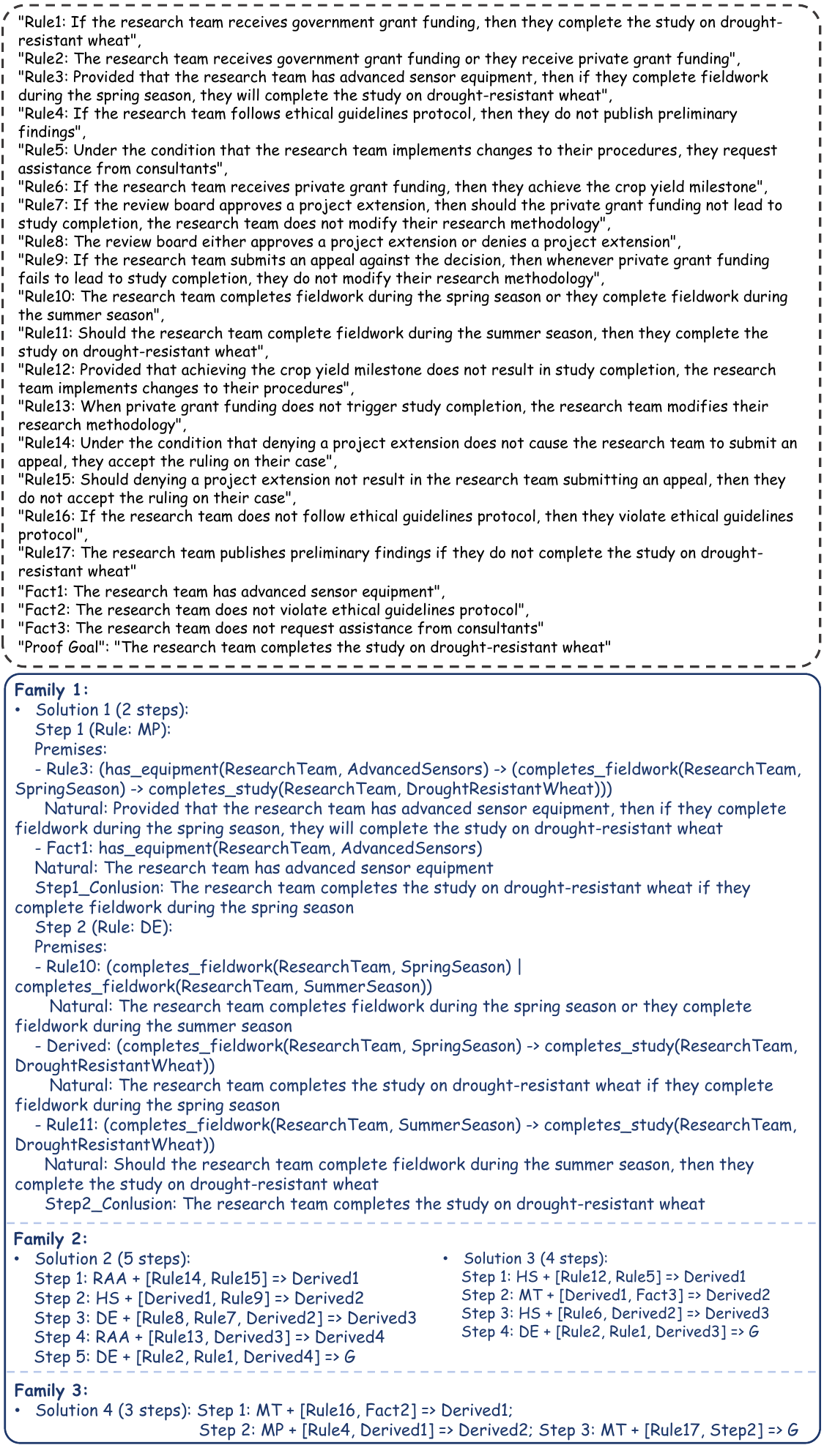}
\vspace{-10px}
\caption{A Small type example in LogicGraph.}
\label{fig:sexam}
\end{figure*}

\begin{figure*}[t]
\centering
\includegraphics[width=0.9\textwidth]{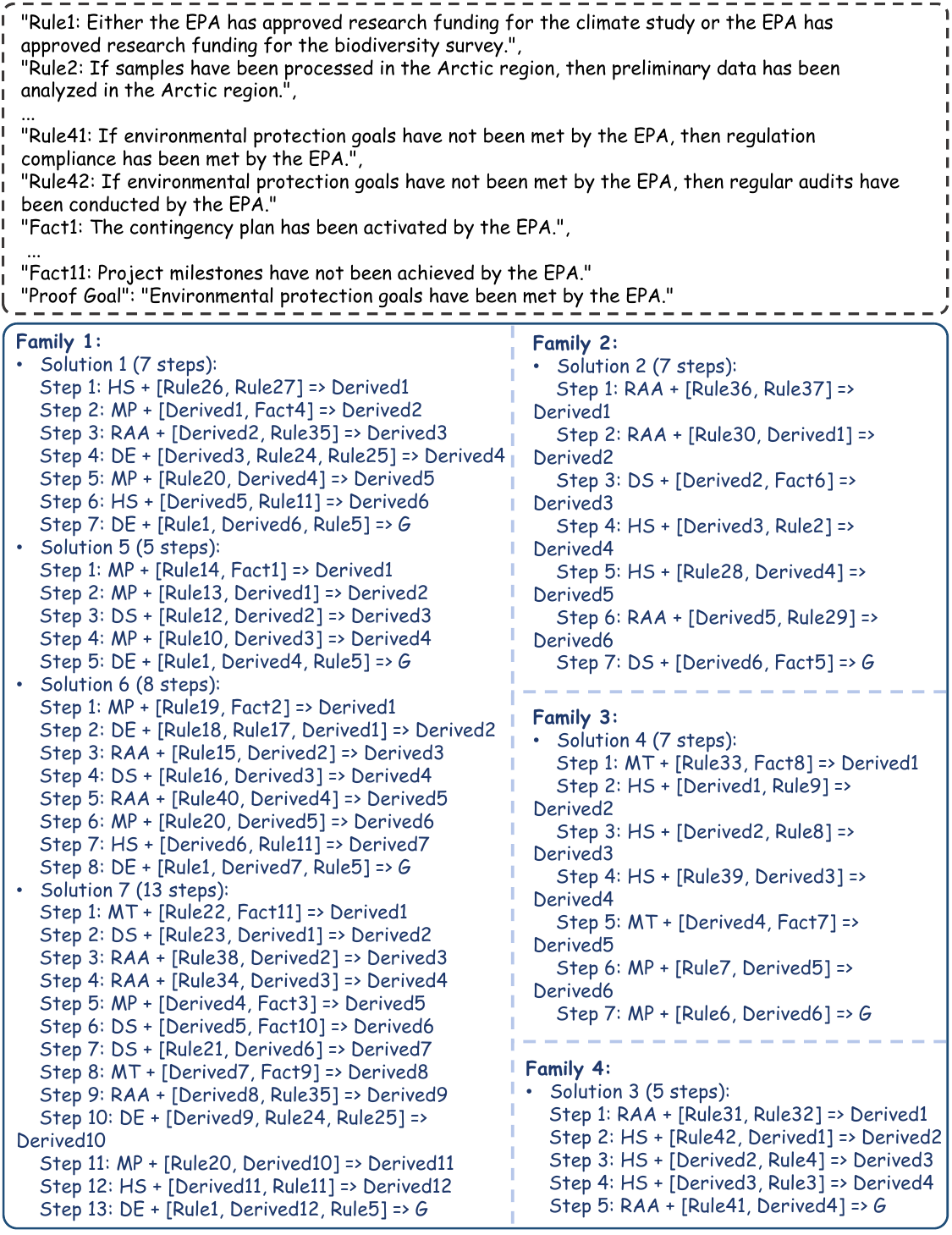}
\caption{A Medium type example in LogicGraph.}
\label{fig:mexam}
\end{figure*}

\begin{figure*}[t]
\centering
\includegraphics[width=0.95\textwidth]{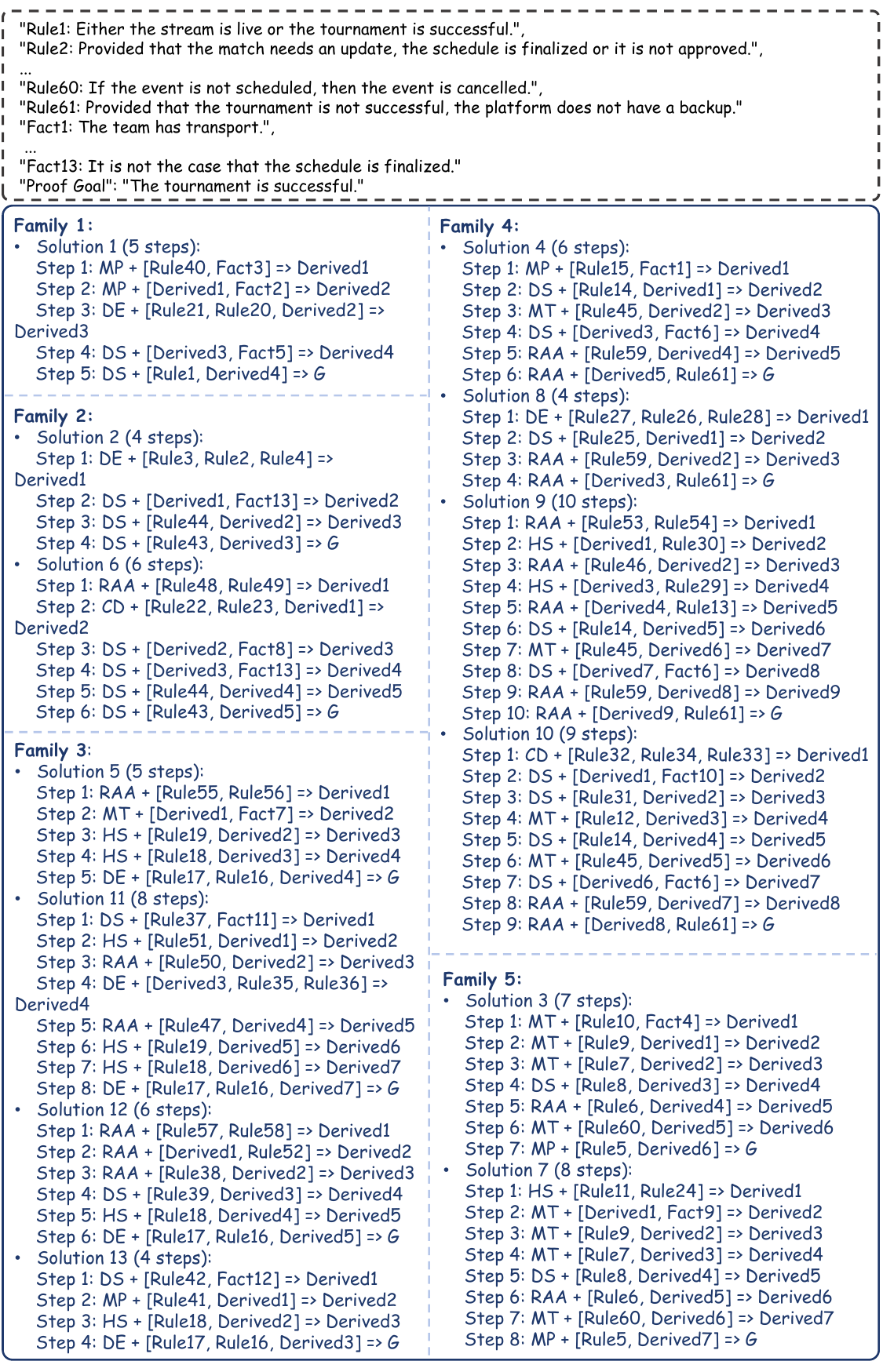}
\caption{A Large type example in LogicGraph.}
\label{fig:lexam}
\end{figure*}

To illustrate the three levels of complexity in our dataset, we present one representative example from each category: Small(Figure~\ref{fig:sexam}), Medium(Figure~\ref{fig:mexam}), and Large(Figure~\ref{fig:lexam}). These cases are categorized by the number of distinct reasoning paths available for the given problem.

Test 5 (Small Case)
This case represents a scenario with a limited number of paths, containing a total of 4 valid solutions. We provide a detailed look at Solution 2, which features comprehensive annotations including symbolic representations, natural language descriptions, and a complete reasoning chain linking premises to conclusions. Notably, Solution 1 and Solution 4 are classified into the same reasoning family (Family 1). This grouping is due to the presence of shared reasoning nodes between the two paths, demonstrating our method of clustering logically similar derivations.

Test 664 (Medium Case)
Representing medium complexity, this test case contains 7 solutions organized into 4 reasoning families. The length of these solutions varies from 5 to 13 steps. The largest family, Family 1, encompasses 4 distinct paths, highlighting the potential for multiple variations within a single line of reasoning.

Test 821 (Hard Case)
This case illustrates a high-complexity scenario with 13 solutions distributed across 5 reasoning families. The solution lengths range from 4 to 10 steps. The abundance of alternative paths and families in this example underscores the depth of the logical problems presented.

These examples collectively demonstrate the quality and depth of our dataset. By capturing a wide range of reasoning complexities—from simple, direct derivations to highly branched logical structures—and providing granular annotations, the dataset offers a rigorous standard for evaluating the reasoning capabilities of Large Language Models.

\section{Error Type Details}
\label{app:error_taxonomy}
\subsection{Detailed Definition}
Table~\ref{tab:app_error_taxonomy} shows detailed definitions and examples of the error taxonomy.
\begin{table*}[t]
\centering
\small
\renewcommand{\arraystretch}{1.4}
\begin{tabularx}{\textwidth}{l p{4.3cm} X p{5cm}}
\toprule \textbf{Error Type} & \textbf{Detailed Description} & \textbf{Concrete Example} \\
\midrule
\multicolumn{3}{c}{\textit{\textbf{A. Semantic Comprehension (Information Processing Layer)}}} \\
\midrule
\textbf{Semantic Misinterpretation} & The model fails to correctly parse the linguistic meaning of a specific fact or rule. This often involves confusing the direction of causality or ignoring negations. & \textit{Context:} ``If A, then B.'' \newline \textit{Model interpretation:} ``If B, then A.'' (Conditional inversion) \newline \textit{Context:} ``John is not eligible.'' \newline \textit{Model interpretation:} ``John is eligible.'' (Negation neglect) \\
\cmidrule{1-3}
 \textbf{Information Omission} & The model completely ignores a critical piece of evidence or a constraint provided in the input context during the reasoning chain. & \textit{Context:} ``John has no medical insurance.'' \newline \textit{Error:} The model concludes John proceeds with a procedure requiring insurance, failing to retrieve the negation fact. \\
\cmidrule{1-3}
\textbf{Fact Hallucination} & The model fabricates premises or assumes external knowledge that contradicts or is absent from the provided context. & \textit{Context:} ``John visits the hospital.'' \newline \textit{Error:} The model adds ``John has cancer'' as a premise for deduction, despite no such mention in the text. \\
\midrule
\multicolumn{3}{c}{\textit{\textbf{B. Logical Execution (Reasoning Layer)}}} \\
\midrule
\textbf{Invalid Deduction} & The derivation step is logically fundamentally flawed. Even if the premises are correct, the conclusion cannot be derived using standard logic (e.g., formal fallacies). & \textit{Premises:} ``Rule: If it rains, the ground is wet.'' + ``Fact: The ground is wet.'' \newline \textit{Conclusion:} ``Therefore, it is raining.'' \newline \textit{Reason:} Fallacy of affirming the consequent. \\
\cmidrule{1-3}
\textbf{Rule Misapplication} & The model selects a rule that exists in the context but applies it to an entity or situation where the rule's preconditions are not met. & \textit{Context:} ``All doctors must be licensed. John is a driver.'' \newline \textit{Error:} Applying the ``doctor licensing'' rule to John, claiming he must have a medical license. \\
\cmidrule{1-3}
\textbf{Insufficient Premise} & The model draws a definitive conclusion based on a subset of premises that are necessary but not sufficient. It fails to aggregate all required conditions. & \textit{Rule:} ``If A and B, then C.'' \newline \textit{Fact:} ``A is true.'' (B is unknown/false). \newline \textit{Error:} Model concludes ``C is true'' solely based on A, ignoring the missing condition B. \\
\bottomrule
\end{tabularx}
\caption{Detailed definitions and examples of the error taxonomy. We provide concrete scenarios to illustrate the distinction between comprehension failures and execution failures.}
\label{tab:app_error_taxonomy}
\end{table*}

\subsection{Error Distribution}
\begin{figure*}[t]
\centering
\includegraphics[width=\textwidth]{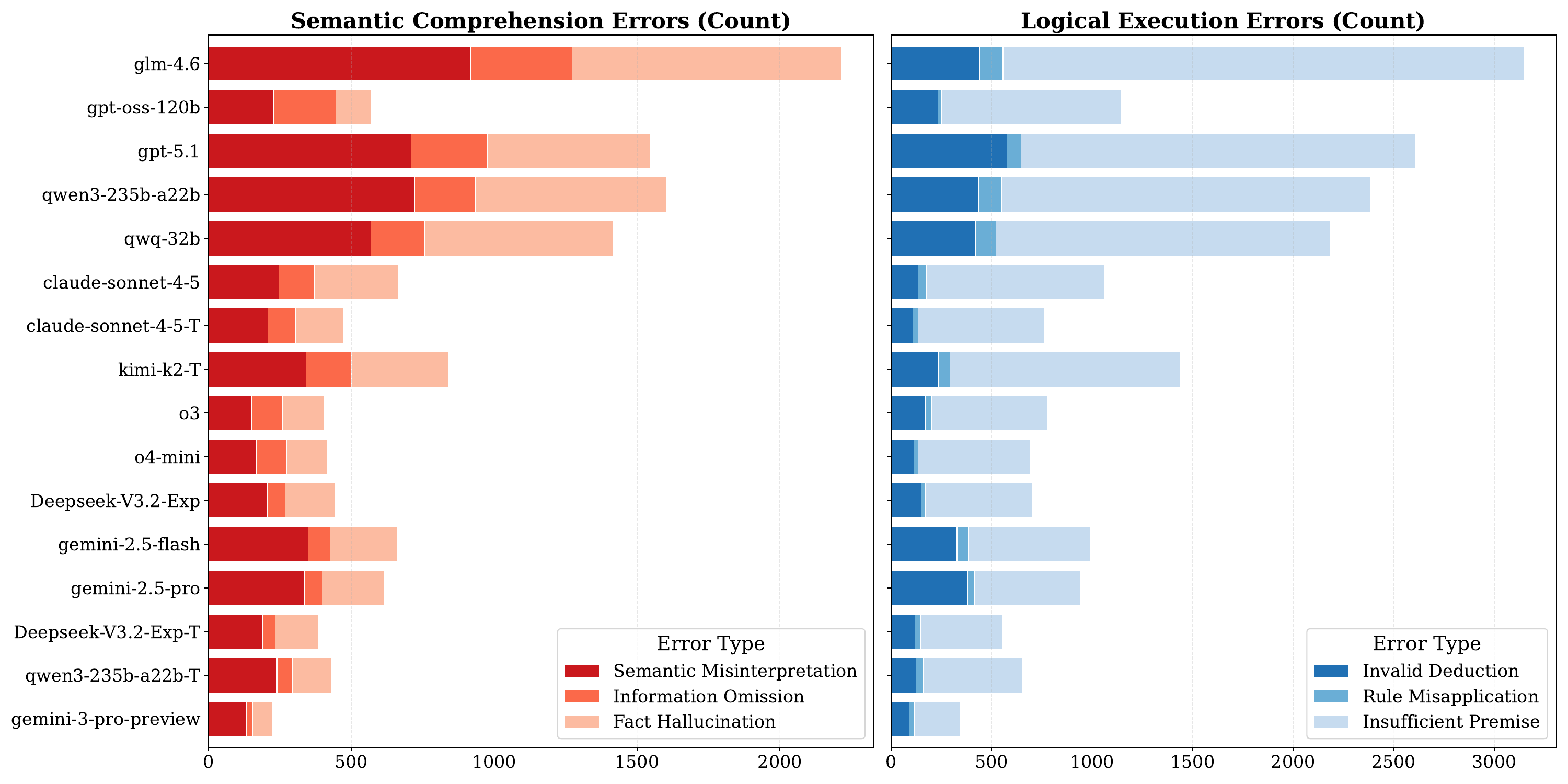}
\caption{Error distribution across models, grouped by semantic comprehension and logical execution.}
\label{fig:eror_dis}
\end{figure*}

Figure~\ref{fig:eror_dis} illustrates the distribution of error types (semantic understanding and logical execution) in models, along with a detailed comparison across different error categories.

\section{Case Study}
\label{app:case}
\subsection{Divergent Metrics Reveal Hidden Reasoning Gaps.}
\label{app:case1}
We present a representative LogicGraph instance with four ground-truth minimal proof paths organized into three distinct reasoning families to qualitatively analyze model behavior beyond aggregate metrics. As shown in Figure~\ref{fig:case}, all evaluated models achieve perfect convergent performance on this case, with Success Rate, Precision, and Shortest Path Finding Rate all reaching 100\%. Under traditional evaluation, these models would therefore appear indistinguishable.
This case illustrates that current LLMs tend to conflate reformulation with
exploration, highlighting the necessity of explicit divergent evaluation.

However, divergent metrics reveal substantial differences in reasoning behavior. Gemini-3-Pro successfully identifies three out of four valid proof paths, covering all reasoning families (Diversity = 75\%, Versatility = 100\%). In contrast, o3, despite producing multiple solutions, repeatedly commits to the same dominant derivation, recovering only a single ground-truth path (Diversity = 25\%, Versatility = 33\%). This indicates early commitment to a high-probability reasoning route, with alternative valid paths remaining unexplored.

A different failure mode is observed for Qwen3-235B-A22B. While the model attempts to enumerate multiple solutions, only two out of six generated paths are logically valid, resulting in a sharp drop in precision. Step-level verification (\ref{app:diag}) shows that invalid solutions often rely on fabricated or insufficient intermediate steps, reflecting result-oriented reasoning rather than structured exploration of the proof space. This highlights that generating more solutions does not necessarily correspond to improved divergent reasoning.

Notably, the reasoning-oriented variant Qwen3-235B-A22B-Thinking achieves full coverage of all ground-truth paths with perfect precision, demonstrating that explicit reasoning optimization can substantially improve both exploration breadth and logical faithfulness in small multi-path settings. Nevertheless, as shown in our large-scale results, such improvements do not fully eliminate the degradation of diversity under increased reasoning depth.

Overall, this case study illustrates that convergent correctness can mask fundamentally different reasoning strategies. Models may reach the correct conclusion while exhibiting pseudo-divergence—superficially diverse outputs that collapse to the same minimal support—or hallucinated exploration driven by result-oriented fabrication. These behaviors underscore the necessity of explicit divergent evaluation for diagnosing genuine multi-path reasoning ability.

\subsection{Diagnosing Logical Fallacies}
\label{app:diag}
\begin{figure*}[t]
\centering
\includegraphics[width=\textwidth]{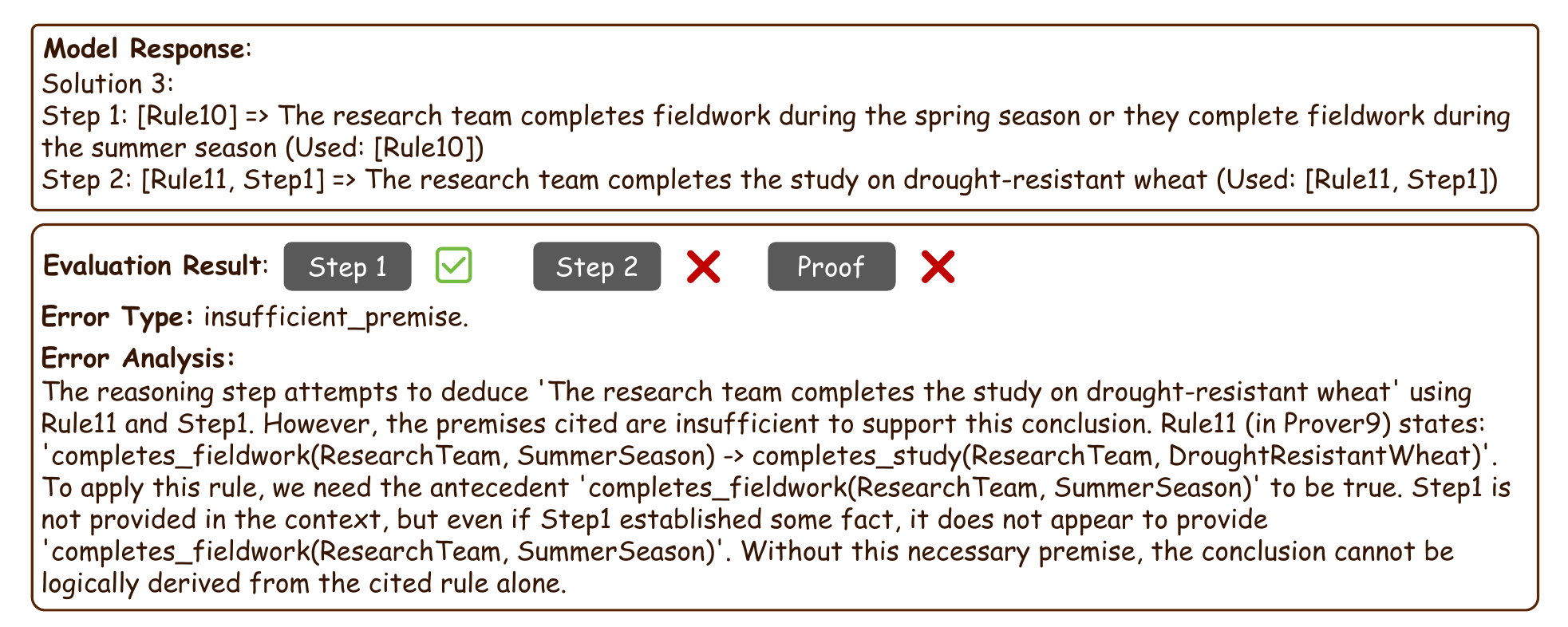}
\caption{A specific example of our model's error analysis.}
\label{fig:error}
\end{figure*}

Extending this analysis to Qwen3-235b-A22b, we observe that high success rates can mask subtle logical flaws in Figure~\ref{fig:error}. While the models arrive at the correct final answer, a granular inspection of the reasoning chains reveals specific failure modes. Notably, our qualitative analysis identifies that the Thinking model frequently succumbs to \textit{insufficient\_premise} errors—making unjustified logical leaps without establishing necessary intermediate lemmas. These findings underscore the necessity of our multi-dimensional evaluation framework for detecting result-oriented fabrications in proof generation.

\subsection{Why Formal Verification Is Necessary for Evaluating Logical Reasoning}
To illustrate the limitations of current LLM-as-a-judge paradigms and to motivate our formal-verification-based evaluation, we provide a qualitative analysis of three representative scenarios observed in the \texttt{test\_11} benchmark. These cases expose both (i) false negatives (rejecting valid but compressed reasoning) and (ii) false positives (accepting formally invalid steps) that frequently occur under direct LLM-based evaluation.

\subsection{False Negatives on Compressed Logical Inference}
\begin{figure*}[t]
\centering
\includegraphics[width=\textwidth]{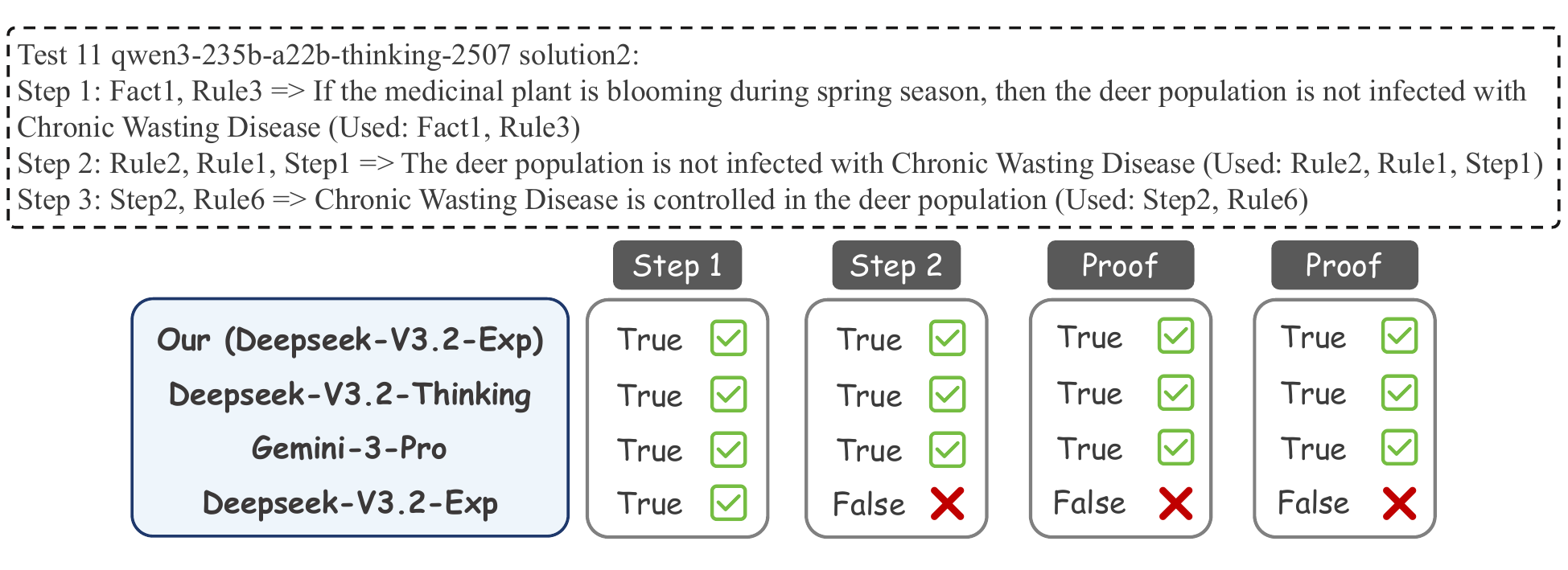}
\caption{A valid complex reasoning step correctly verified by our formal method but incorrectly accepted by an LLM-based judge (using Deepseek-V3.2-Exp directly as an LLM judge).}
\label{fig:eva_exam1}
\end{figure*}

LLM-based evaluators often struggle when multiple logical operations are compressed into a single step, which can lead to false negatives.
Figure~\ref{fig:eva_exam1} shows a solution produced by qwen3-235b-a22b-thinking. In Step~2, the model effectively applies a CD form, and the  premises correspond to:
\begin{align*}
(\text{Rule 2})\;& \mathsf{ResearchTeam} \lor \mathsf{MedicinalPlant} \\
(\text{Rule 1})\;& \mathsf{ResearchTeam} \rightarrow \neg \mathsf{Infected} \\
(\text{Step 1})\;& \mathsf{MedicinalPlant} \rightarrow \neg \mathsf{Infected}
\end{align*}

\noindent
Our verification pipeline (\texttt{Deepseek-V3.2-Exp} with Prover9) successfully proves the target and validates Step~2. However, when using \texttt{Deepseek-V3.2-Exp} directly as an LLM judge (i.e., without formal proof checking), the same step is incorrectly flagged as invalid. This indicates that standard LLM evaluators may lack the robustness to assess logically correct but highly compressed inferences, inadvertently penalizing models that employ higher-efficiency reasoning patterns.

\subsection{False Positives and the Gap-Filling Bias of LLM Judges}
A more severe failure mode of LLM-as-a-judge is the tendency to over-credit incomplete derivations by implicitly supplying missing premises (i.e., ``filling in the gaps''). We compare two \texttt{test\_11} solutions to demonstrate this issue.

\begin{figure*}[t]
\centering
\includegraphics[width=\textwidth]{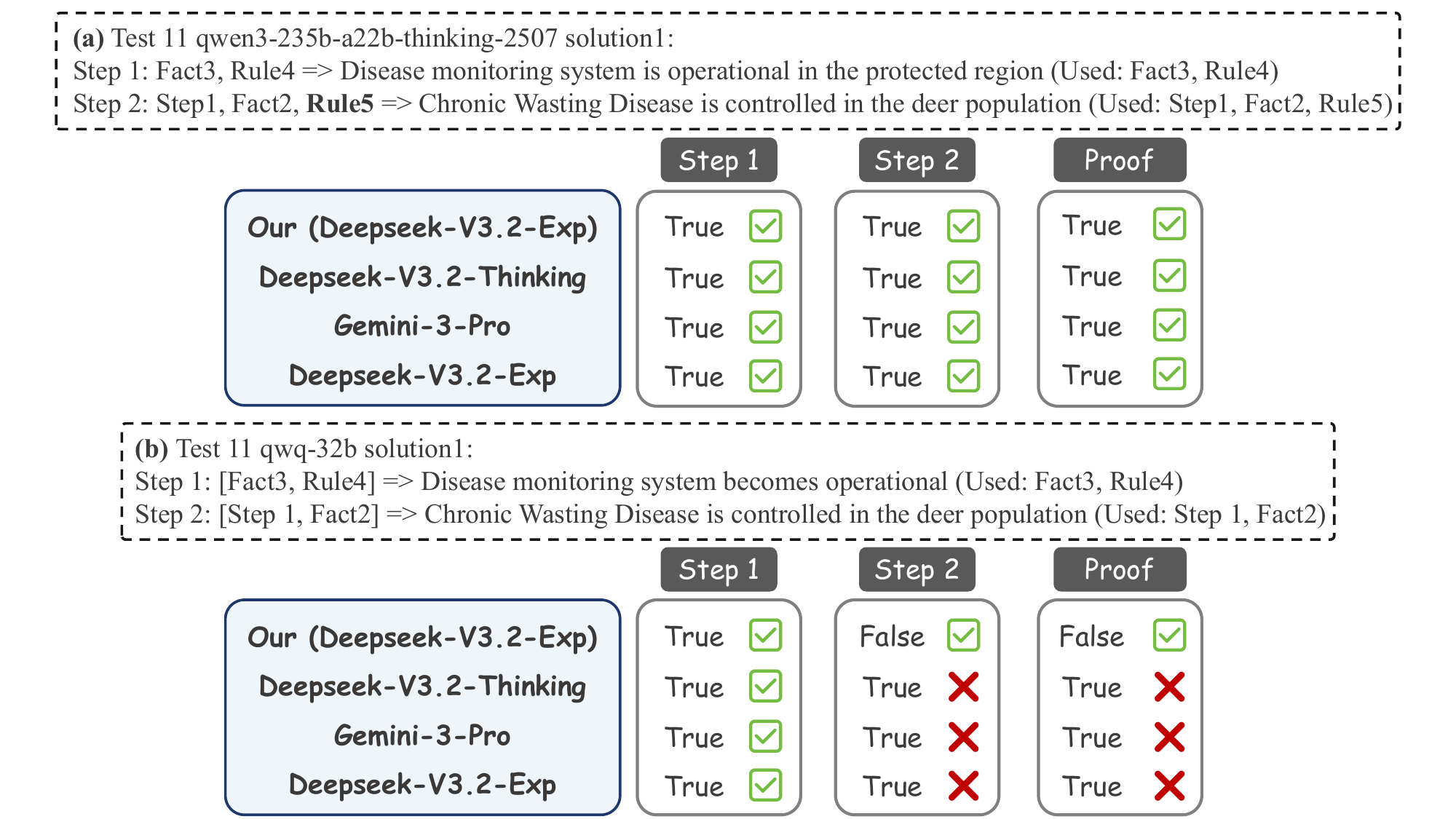}
\caption{Comparison of evaluation robustness. (a) A fully valid solution. (b) An invalid solution with missing premises (Rule~5), correctly rejected by our method but incorrectly accepted by LLM judges.}
\label{fig:eva_exam2}
\end{figure*}

Figure~\ref{fig:eva_exam2}(a) presents a correct solution from \texttt{qwen3-235b-a22b-thinking}. In Step~2, the derivation explicitly cites the necessary bridge rule:
\[
(\text{Rule 5})\quad \mathsf{Operational} \land \mathsf{Dispensers} \rightarrow \mathsf{Controlled},
\]
together with the required factual premises. All evaluation methods correctly identify this solution as valid.

Figure~\ref{fig:eva_exam2}(b) shows a flawed solution from \texttt{qwq-32b}. In Step~2, the model attempts to derive the same conclusion using only:
\[
(\text{Step 1})\ \mathsf{Operational}
\qquad
(\text{Fact 2})\ \mathsf{Dispensers},
\]
while omitting Rule~5. Formally, without the implication rule we have
\[
\mathsf{Operational} \land \mathsf{Dispensers} \nvdash \mathsf{Controlled},
\]
so the step is a non sequitur.

\noindent

Our formal verifier correctly rejects this step (\texttt{False}). In contrast, the other LLM-based evaluators mark it as valid (\texttt{True}). This highlights a systematic bias: LLM judges tend to prioritize semantic plausibility over logical entailment, and may implicitly reconstruct missing rules from their own priors or context. As a result, conventional LLM-only evaluation can inflate scores for ``nearly correct'' but formally invalid proofs, making it unreliable for measuring strict logical reasoning ability.

\section{Prompt for Dataset Generation}
\label{app:generation}
\begin{figure*}[t]
    \centering
    \includegraphics[width=0.95\textwidth]{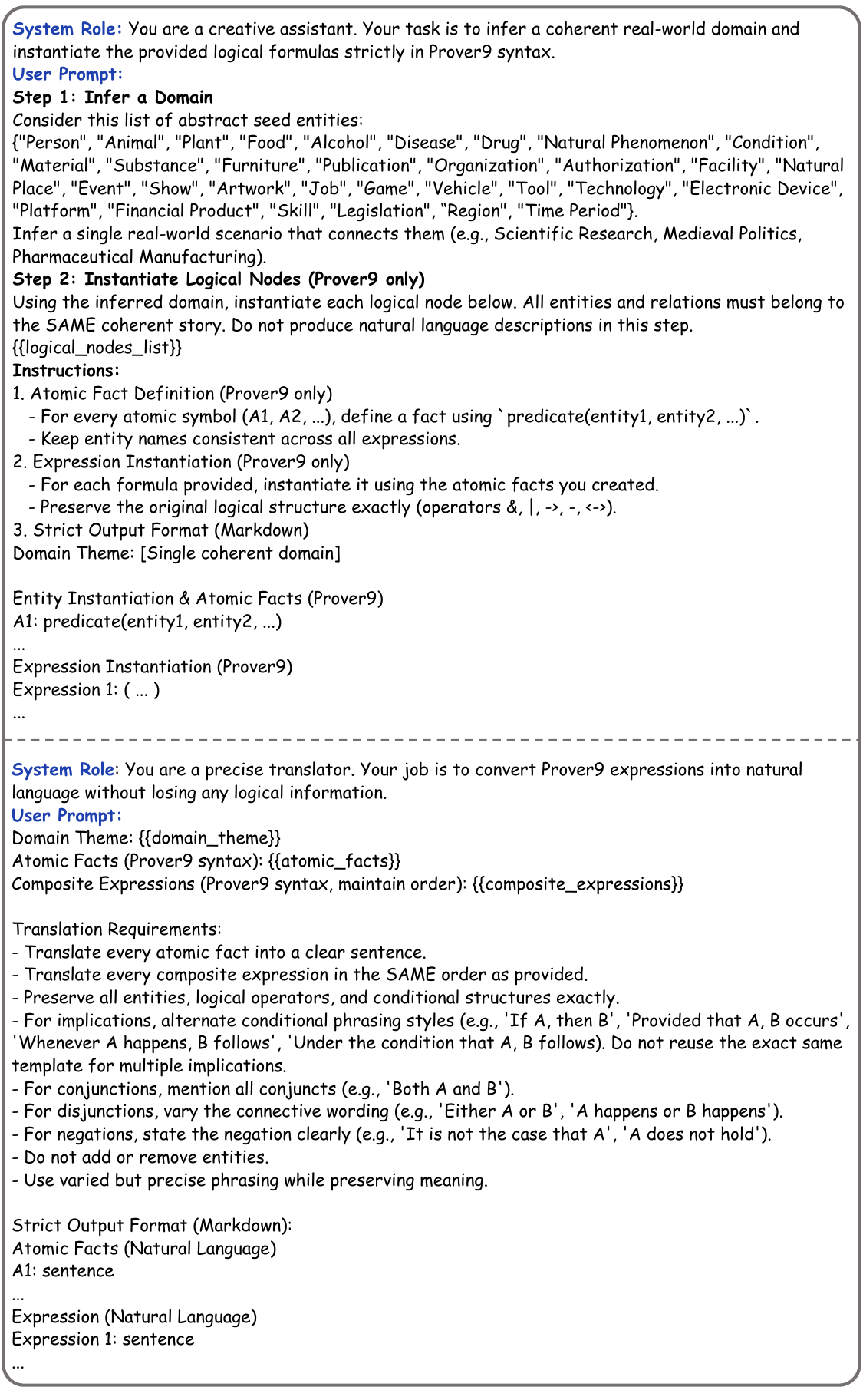}
    \vspace{-10px}
    \caption{Prompt we use during the generation process.}
    \label{fig:app_generation}
\end{figure*}

Figure \ref{fig:app_generation} shows the prompt we used to infer a coherent real-world domain from a set of abstract entities and instantiate logical formulas into Prover9 syntax. Below is the prompt we used to translate the instantiated Prover9 expressions into natural language to form the final dataset.

\section{Prompt for Model Evaluation}
\label{app:prompts}
\begin{figure*}[t]
    \centering
    \includegraphics[width=0.95\textwidth]{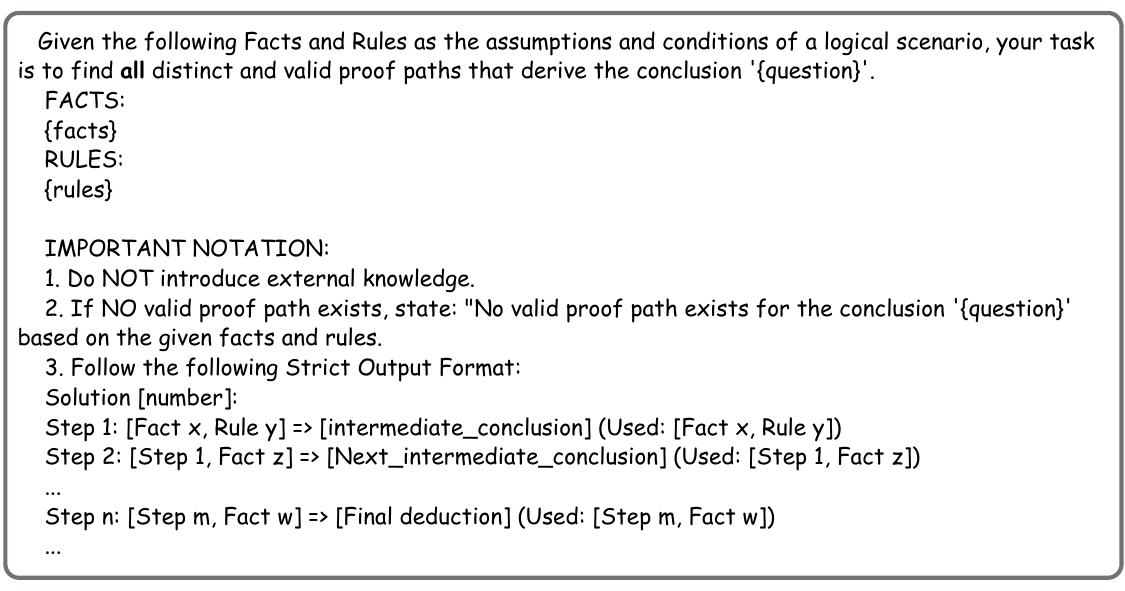}
    \vspace{-10px}
    \caption{Prompt we use during the test process.}
    \label{fig:app_test}
\end{figure*}

Figure \ref{fig:app_test} shows the prompt we used to test the Large Language Models (LLMs) on the multi-path reasoning task.

\section{Prompt for Model Performance Evaluation}
We present the condensed prompts used in our evaluation pipeline. Dynamic content injected at runtime is denoted by brackets (e.g., {{CONTEXT}}).

\begin{figure*}[t]
    \centering
    \includegraphics[width=0.95\textwidth]{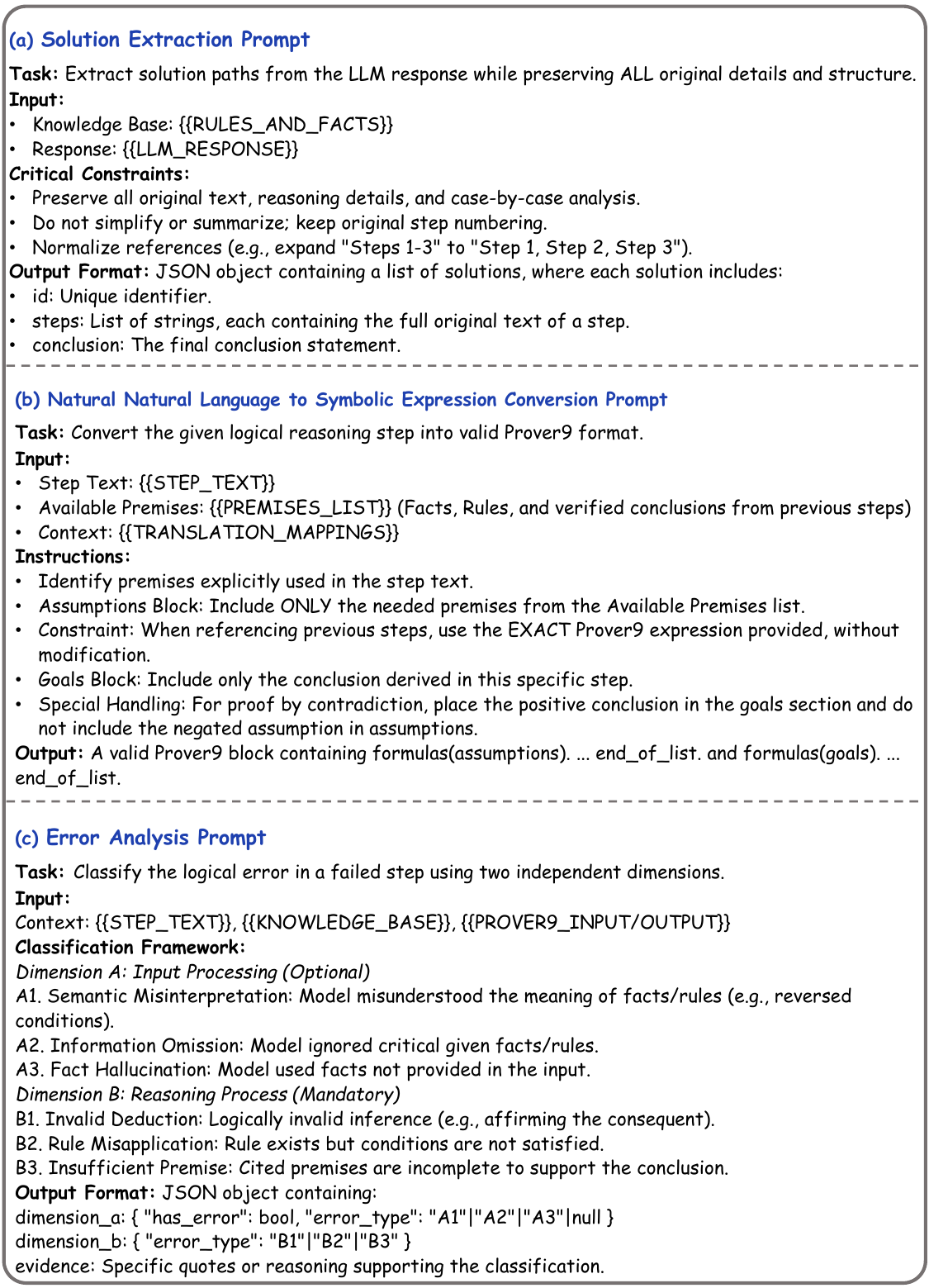}
    \vspace{-5px}
    \caption{Prompt template we use during the performance evaluation process.}
    \label{fig:evaluator}
\end{figure*}

\label{app:evaluator}
Figure \ref{fig:evaluator} shows the prompt we used by the evaluator agent to parse the LLM's raw response into structured JSON for automated verification.

\section{Details of human annotators}
For data qualitative evaluation, we engaged four graduate students (including both PhD and Master's students) from engineering disciplines who are also co-authors of this paper. All annotators possessed strong backgrounds in formal logic and critical reasoning, making them well-qualified for this task. Since the annotators were co-authors actively involved in the research, no formal recruitment process or compensation was required, and they were fully aware of how the data would be used in the study. 
The annotation process focused solely on logical validity and natural language quality evaluation and did not involve collecting any personal identifying information or expose annotators to any risks.
As this research involved co-authors analyzing academic content rather than external human subjects, it was determined to be exempt from formal ethics review board approval. 
The annotation work was conducted as part of regular academic research activities within our institution.  
No protected or sensitive demographic information was collected or used in this research.

\section{Details of Ai Assistants In Research Or Writing}
Artificial Intelligence tools were utilized in two distinct capacities during this research:

\paragraph{Dataset Generation.}
To construct the multi-solution logic benchmark, we utilized LLMs (specifically DeepSeek-V3 via API) to translate symbolic logic templates into natural language narratives. This process involved converting formal logical rules (e.g., implications, disjunctions) into coherent scenarios across various domains such as cybersecurity, law, and medical diagnosis. The generated text was subsequently verified by human annotators to ensure clarity and fidelity to the underlying logic.

\paragraph{Research and Writing Assistance.}
We utilized GitHub Copilot (powered by Gemini 3 Pro) as a coding assistant to help develop the evaluation scripts and analyze the experimental data. Additionally, AI assistance

\section{Details of Computational Experiment}
Our computational experiments evaluated a diverse set of Large Language Models (LLMs) for logical reasoning, covering both proprietary API-based systems and open-weight releases.

\paragraph{Models Evaluated.}
We evaluated proprietary models accessed via APIs, including GLM-4.6, GPT-5.1, Claude-Sonnet-4.5, o3/o4-mini, and Gemini 2.5-Flash, Gemini-2.5-Pro, Gemini-3-Pro-Preview,. We also evaluated open-weight models, including GPT-OSS-120B, Qwen3-235B-A22B (and its Thinking variant)~\cite{qwen3}, QwQ-32B~\cite{qwq_32b}, and DeepSeek-V3.2-Exp (and its Thinking variant)~\cite{deepseek_v32exp}. For all models, we adopted a unified prompting protocol that encourages the model to produce as many independent and verifiable solution paths as possible under a fixed structured answer template; prompt templates are provided in Appendix~\ref{app:prompts}.

\paragraph{Deployment and Access.}
All proprietary models were queried through their official APIs. Among the open-weight models, only QwQ-32B was deployed locally for inference; it was tested on a machine equipped with two NVIDIA A100 GPUs. All other models were accessed via APIs.

\paragraph{Evaluation Framework.}
We implemented an automated evaluation pipeline to assess model outputs. In particular, we performed formal verification using the automated theorem prover \texttt{Prover9} to check the logical validity of the generated derivations. Only logically valid steps, as verified by \texttt{Prover9}, were counted toward the final evaluation metrics.

\end{document}